\documentclass[runningheads]{llncs}

\usepackage{eccv}

\usepackage{eccvabbrv}
\usepackage{amsmath,amsfonts,bm}

\usepackage{graphicx}
\usepackage{booktabs}
\usepackage{tabularx} 
\usepackage{multirow} 
\usepackage{wrapfig}
\usepackage{makecell}
\usepackage{colortbl}
\usepackage{pifont}
\usepackage{xcolor}
\usepackage{tabularray}
\usepackage{algorithmic}
\usepackage{cuted}    
\usepackage{dsfont}

\usepackage[accsupp]{axessibility}  

\newcommand{\cmark}{\textcolor{green!80!black}{\ding{51}}}
\newcommand{\xmark}{\textcolor{red}{\ding{55}}}
\newcommand{\supp}[1]{{\color{blue}  #1}}

\newcommand{\venue}[1]{{$_{\text{#1}}$}}
\makeatletter
\newcommand{\enableindentfirst}{%
  \let\@afterindentfalse\@afterindenttrue
  \@afterindenttrue
}
\makeatother

\def\vb{{\bm{b}}}

\def\vq{{\bm{q}}}

\def\vt{{\bm{t}}}

\def\vv{{\bm{v}}}

\def\vy{{\bm{y}}}

\def\mA{{\bm{A}}}

\def\mV{{\bm{V}}}
\def\mW{{\bm{W}}}

\usepackage{hyperref}

\usepackage{orcidlink}

\begin{document}

\title{Controlling Embedding Spaces with Text-Conditioned Transformations} 


\author{Joseph Fioresi\inst{1}\thanks{Majority of work done as an intern at Adobe Research}\orcidlink{0000-0001-6898-3689} \and
Fabian Caba Heilbron\inst{2}\orcidlink{0000-0002-3129-1985} \and
Pankaj Nathani\inst{2} \and \\ Mubarak Shah\inst{1}\orcidlink{0000-0001-6172-5572} \and Kushal Kafle\inst{2}\orcidlink{0000-0002-0847-7861}}

\authorrunning{J.~Fioresi et al.}

\institute{Institute of Artificial Intelligence, University of Central Florida, USA \and
Adobe Research, USA\\
\email{joseph.fioresi@ucf.edu}, \email{\{caba,pankajn,kkafle\}@adobe.com}, \email{shah@crcv.ucf.edu}
\\ \url{https://joefioresi718.github.io/ControlEmbed_webpage/}
}

\maketitle

\begin{abstract}
  Multimodal embedding spaces in models like CLIP enable powerful capabilities such as semantic similarity retrieval and cross-modal zero-shot classification. These embeddings compress high-level semantics into a single vector, which comes at the cost of primarily expressing a dominant semantics like main object while suppressing other important attributes such as camera angle or color tone. We propose a text-conditioned transformation of visual embeddings that makes such attributes explicitly accessible. Given a natural language description of an attribute category (e.g., “color” or “art style”), a network generates an affine transformation that emphasizes the specified attribute. Conditioning on text enables it to learn many attributes simultaneously, accessing them at inference time through an intuitive interface. The network is trained to align transformed embeddings with the frozen latent space, enabling retrieval using existing large-scale embeddings without any re-encoding. When applied to a full set, the same mechanism transforms the latent space for attribute disentanglement tasks such as multi-clustering. By operating directly in latent space, our method provides a unified and efficient framework for controlling embedding spaces, demonstrating state-of-the-art performance across both attribute-based retrieval and multi-attribute organization tasks with near-zero inference cost.
  \keywords{Vision-Language Models \and Embedding Spaces}
\end{abstract}

\begin{figure}[h]
    \centering
    \includegraphics[width=\linewidth]{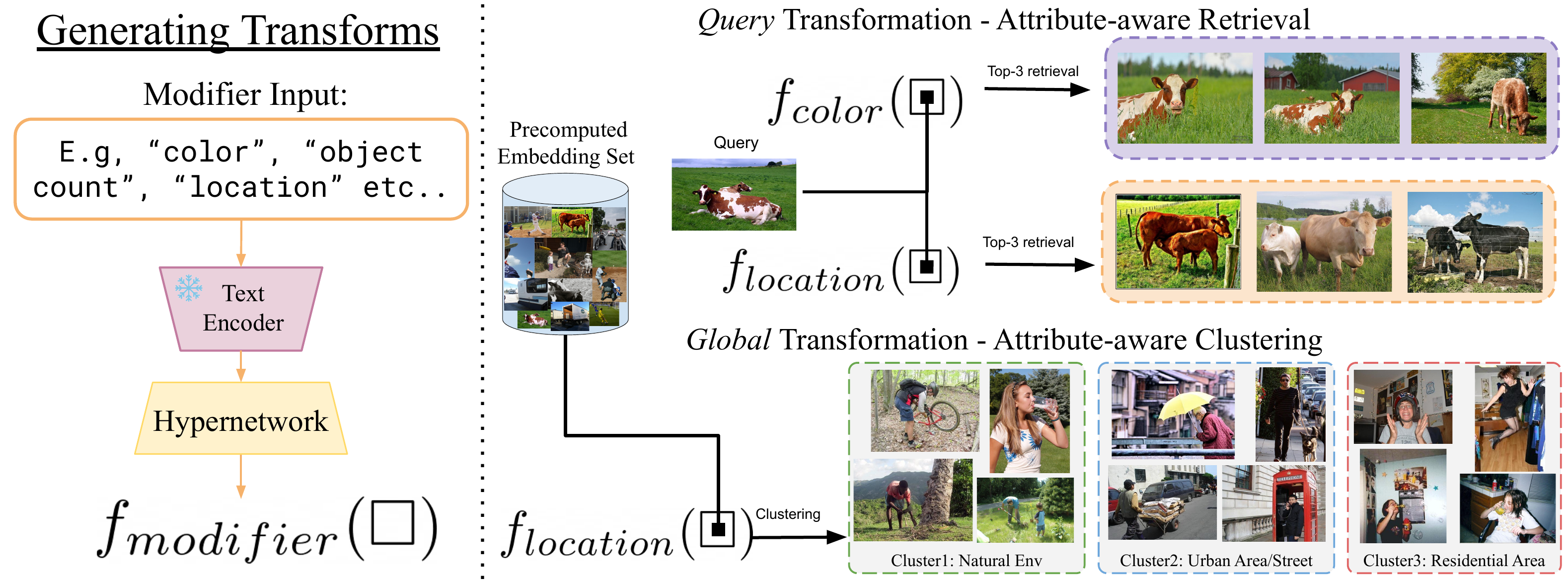}
    \caption{Overview of our text-conditioned generation of embedding-space transformations. (Left) We propose a hypernetwork that maps natural-language attribute descriptions to a modification function. Applying this transformation to a query embedding produces an attribute-aware representation that emphasizes the requested property. (Right) By controlling the embedding space with attribute transforms, the same mechanism enables both attribute-aware retrieval and attribute-specific clustering.
 }
    \label{fig:teaser}
\end{figure}

\section{Introduction}
\label{sec:intro}

Large-scale vision-language models like CLIP~\cite{radford2021learning} and SigLIP~\cite{zhai2023sigmoid} have become ubiquitous in computer vision, learning powerful multimodal embeddings from web-scale data. Such embeddings enable zero-shot task transfer, cross-modal retrieval, and semantic similarity search, and are increasingly deployed in practice. However, their versatility comes from compressing diverse visual concepts---objects, attributes, styles, and contexts---into a single high-dimensional space. As a result, important attributes such as color, camera angle, or mood are encoded only implicitly, entangled with more dominant object-level semantics. This entanglement limits the kinds of similarity queries that can be expressed. When retrieving images from a reference image with CLIP, we can neither predict nor control whether the similarity will be based on the main subject, style, or other attributes. For example, suppose a reference image is a pastel-colored watercolor painting of a Corgi as viewed from a low angle. In this case, CLIP can retrieve images that look similar, but the specific properties are nebulous~\cite{tong2024eyes}. Will a photograph of a Corgi against a plain background rank higher or lower than a similarly styled painting of a Labrador viewed from a similar angle? Not only can we not readily know the answer, but we also cannot easily specify our preference. 

This limitation restricts the feasibility of \textit{attribute-aware} retrieval, where users search based on specific attributes rather than broad semantic categories. For instance, given a large collection of photos, one user might want to retrieve images with similar composition or lighting, while another might care more about color or artistic style. Yet, because such attributes are entangled within the latent space, a single query embedding cannot easily emphasize them. This challenge extends beyond retrieval to other tasks such as \textit{multi-clustering}, where different users may wish to group the same media collection by different attributes. Together, these limitations highlight the need for controllable transformations of embedding spaces that adapt dynamically to user-specified attributes.

Existing methods for attribute disentanglement often rely on one of two major strategies: (1) finetuning the encoder to explicitly separate desired attributes~\cite{wang2023improved, gao2024clip, yao2024customized, yao2024multi}; or (2) scaling up inference-time requirements by composing knowledge from large VLMs and LLMs~\cite{kwon2023image, liu2025organizing, luo2025llm}. The first approach enables efficient inference once trained, but demands costly finetuning and re-embedding. This causes linear memory footprint scaling and sacrifices the generalization of the original model. The second requires no retraining, but introduces high inference latency and limited scalability for real-time applications. 

In this paper, we propose a novel approach to address the above limitations. At the core of our approach is a simple but powerful idea: instead of retraining the encoder or designing separate attribute-specific heads, we learn to reshape the latent space itself through a text-conditioned transformation. Concretely, we train an MLP-based hypernetwork that takes as input a text embedding representing an attribute category (e.g., ``color'' or ``style'') and outputs a single global affine transformation. This transform acts as a learned operator that modifies the existing embedding space so that the specified attribute becomes more salient. Figure~\ref{fig:teaser} demonstrates this idea, showcasing its use in two complementary applications: a \textit{query} transformation, where only the query embedding is modified to emphasize a desired attribute for attribute-based retrieval from a static embedding index; and a \textit{global} transformation, which transforms all embeddings for downstream analysis such as unsupervised clustering. Because the encoder remains frozen and the hypernetwork operates directly on latent embeddings, inference is near-instant and the method can be seamlessly integrated into existing multimodal pipelines.

Our main contributions are summarized as follows:
\begin{itemize}
    \item We introduce a \textbf{text-conditioned hypernetwork} that generates transformations of visual embeddings, enabling controlled emphasis of diverse attributes such as color, mood, and camera angle. Alignment with base embeddings enables a \textit{query} mode for attribute-aware retrieval.
    \item These transformations can be applied to disentangle entire embedding sets in a \textit{global} setting. This induces an \textbf{attribute-specific similarity metric}, where the pairwise distances reflect closeness along the chosen category axis.
    \item Extensive experiments demonstrate that our lightweight transformations achieve performance matching or surpassing existing approaches across multiple attribute control tasks with minimal computational cost. A single hypernetwork is trained across \textbf{all} attributes/datasets simultaneously, for a total coverage of \textbf{19} different attributes with \textbf{187} unique subclasses.
\end{itemize}

\section{Related Work}
\label{sec:rel_works}

\noindent\textbf{Foundation embedding models.} 
Foundation models for vision-text embedding have demonstrated that high-dimensional representations can capture rich semantic relationships from large-scale text data~\cite{mikolov2013efficient, pennington2014glove}. This principle has been extended to multimodal contexts by models such as CLIP~\cite{radford2021learning}, ALIGN~\cite{jia2021scaling}, and SigLIP~\cite{zhai2023sigmoid}, which learn a joint embedding space for images and text. Subsequent work has further scaled and refined such embeddings to improve representation quality and generalization~\cite{tschannen2025siglip, oquab2023dinov2, kusupati2022matryoshka, venkataramanan2025franca, Swetha_2023_ICCV}. Our approach operates on these frozen embeddings, allowing seamless integration into existing pipelines without recomputing features, and avoids fine-tuning the encoder, thereby mitigating forgetting or performance degradation.

Powerful multimodal embeddings have enabled applications such as semantic image retrieval~\cite{noh2017large, simeoni2019local, cao2020unifying, zhang2023learning} and zero-shot classification~\cite{radford2021learning, menon2022visual, zhou2022conditional, wortsman2022robust, Swetha_Xformer_ECCV2024, swetha2026smpro}. Many works proposed lightweight task adaptions of such models to improve downstream performance~\cite{hu2022lora, khattakMaPLe}. Composed or conditional image retrieval methods extend these capabilities by allowing a user-specified text modification to retrieve images along different semantic axes~\cite{vo2019composing, liu2021image, baldrati2022conditioned, baldrati2023zero, tian2023fashion, liu2023candidate, agnolucci2025isearle, gupta2025play, wang2025generative, wang2026generating}. In contrast, our method focuses on \textit{controlling embeddings} according to a text query, either by transforming the query alone for retrieval or by transforming the full embedding space, rather than retrieving a specific modified image.

\noindent\textbf{Attribute disentanglement/attribute-aware embeddings.} 
While foundation models like CLIP and ALIGN provide rich multimodal embeddings, fine-grained attributes such as color, pose, and lighting are often entangled with higher-level semantics~\cite{tong2024eyes}. Recent approaches have sought to disentangle visual embeddings into separate attribute and object components. For example,~\cite{saini2022disentangling, hou2021learning} learn disentangled representations to support compositional reasoning and fashion-based retrieval, respectively, but require retraining, which can be computationally expensive. Other works adopt lightweight approaches using frozen embeddings~\cite{wei2023elite, valevski2023face0}, but focus primarily on personalization for text-to-image generation. Another line of research aims to learn interpretable attribute bits for hash-code-based retrieval~\cite{cui2020exchnet, wei20212, shen2022semicon, lu2023attributes, wei2023attribute, chen2024characteristics, wang2025learning}. In contrast, our method proposes a hypernetwork to generate controlled transformations of embeddings, enabling robust disentanglement without retraining the base encoder.

\noindent\textbf{Multi-clustering methods.}
Multi-clustering aims to cluster the same dataset in various meaningful ways. Traditional methods often require careful feature selection and design~\cite{hu2017finding, veit2017conditional, wei2020multi, miklautz2020deep, ren2022diversified, yao2023augdmc}. Recent works like Multi-Map~\cite{yao2024multi} and Multi-Sub~\cite{yao2024customized} leverage powerful pretrained models like CLIP but fine-tune the model on each user query, optimizing clustering-specific losses that can disturb the natural relational structure. Another branch of methods utilizes large multimodal models (LMMs) such as LLaVA~\cite{liu2023visual} to caption images, then employs large language models (LLMs) for iterative clustering based on text-based user inputs~\cite{kwon2023image, luo2025llm, liu2025organizing}. While these methods avoid retraining, they introduce significant inference costs. In contrast, our method learns to generate a text-conditioned projection that operates on frozen encoder features, offering a scalable and efficient solution for multi-clustering tasks.

\section{Method}
\label{sec:method}

The main objective of this work is to construct meaningful, text-conditioned transformations of foundation model embedding spaces. Our approach learns a \textit{single} hypernetwork that maps a natural language input (e.g., ``color'' or ``camera angle'') to the parameters of an affine transformation, reshaping the embedding space to emphasize the queried attribute. This transformation can be applied either to (1) the retrieval query only for efficient attribute-aware retrieval without changing the pre-extracted visual features for the whole gallery, or (2) the full embedding set for global tasks such as clustering.

\begin{figure}
    \centering
    \includegraphics[width=0.9\linewidth]{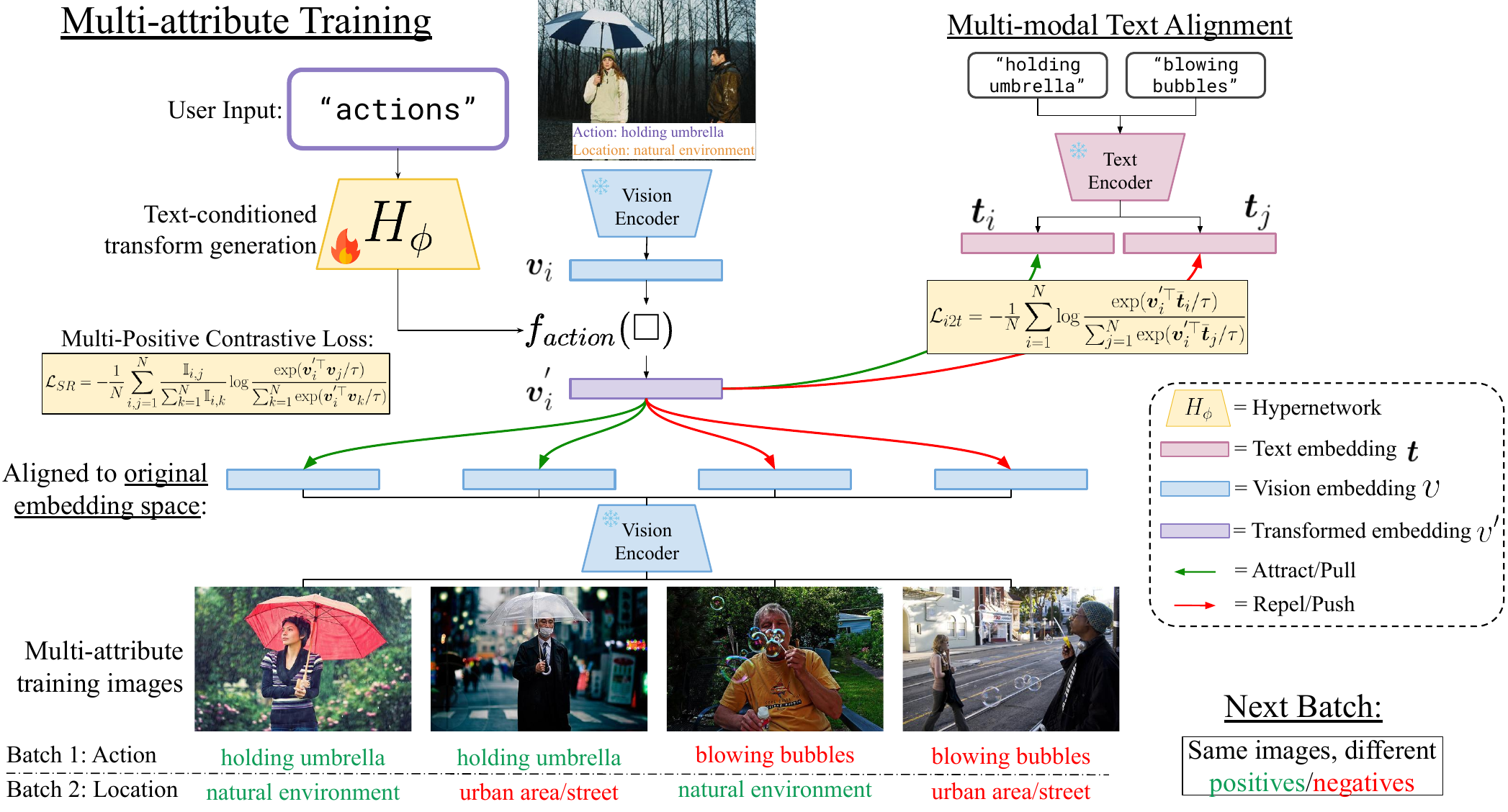}
    \caption{Overview of our multi-attribute training workflow. An attribute category is sampled (e.g., “action” or “location”) and passed through the proposed hypernetwork to generate a transformation function, which is applied to a query embedding to produce attribute-focused representations aligned with base embeddings of images sharing the same category attribute label via a multi-positive contrastive loss. An additional multimodal guidance loss further aligns transformed features with the text embedding of the corresponding attribute label. The process iterates over the full set of valid attributes, where the set of positive/negative samples varies based on selected category.}
    \label{fig:workflow}
\end{figure}

\subsection{Notation}
We denote the vision and text encoders as $f_v$ and $f_t$, respectively. Given an image $I$ and a text input $T$, the encoders produce modality-specific embeddings:
\[
\vv = f_v(I), \quad \vt = f_t(T),
\]
where $\vv, \vt \in \mathbb{R}^{D}$.
Unless otherwise specified, we use the CLIP ViT-L/14 model~\cite{radford2021learning} with frozen  weights. All operations are performed directly in the embedding space, eliminating the need for finetuning the base encoders or feature re-extraction.

We consider a dataset $\mathcal{D} = \{ (I_i, \vy_i) \}_{i=1}^N$, where each image $I_i$ is annotated with labels $\vy_i$ across $M$ semantic taxonomies. Each category $\mathcal{C}^{(m)}$ (e.g., ``color'') defines a set of $K_m$ mutually exclusive classes $\{ c_1^{(m)}, \dots, c_{K_m}^{(m)} \}$.

\subsection{Text-Conditioned Affine Transformation Hypernetwork}

We propose a hypernetwork $H_{\phi}$ that generates an affine transformation conditioned on a textual query. Given a category text embedding $\vt$, the hypernetwork outputs the parameters of an affine transformation:
\begin{equation}
    \mW,\vb = H_{\phi}(\vt),
\end{equation}
where $\mW \in \mathbb{R}^{D \times D}$ and $\vb \in \mathbb{R}^{D}$. Each image embedding $\vv$ is then projected as:
\begin{equation}
    \vv' = \vv \mW^T + \vb.
    \label{eq:affine_transform}
\end{equation}
This can be applied to a single query embedding for attribute-aware retrieval in the \textit{query} setting, or to the full embedding set $\mV$ in the \textit{global} setting to create a new embedding space $\mV'$ specialized to the input attribute. Note that the conditioning signal does not fundamentally require text embeddings, it is chosen as a natural interface choice for specifying attributes.

\noindent\textbf{Query-based Training Objective.}
The hypernetwork is trained to generate transformations that reorganize embeddings based on a user-specified category while maintaining consistency with the foundation model’s original embedding structure. An attribute category is sampled for each batch, defining the label set for the loss objectives. Diverse sampling results in a single hypernetwork capable of optimizing for multiple attributes simultaneously, shown in Figure~\ref{fig:workflow}. We use a multi-positive contrastive loss based on StableRep~\cite{tian2023stablerep} to encourage transformed samples to be near the base embeddings from the same subclass:

\begin{equation}
\label{eq:stablerep}
\mathcal{L}_{SR}=-\frac{1}{N} \sum_{i,j=1}^N \frac{\mathds{1}_{i,j}}{\sum_{k=1}^N \mathds{1}_{i,k}} \log \frac{\exp(\vv_{i}^{'\top} \vv_{j} / \tau)}{\sum_{k=1}^N \exp(\vv_{i}^{'\top} \vv_{k} / \tau)},
\end{equation}
where $\vv^{'}_i \in \mV^{'}, \vv_j \in \mV$ are $\ell_2$-normalized feature embeddings, $\tau$ is a temperature parameter, $N$ is the batch size, and $\mathds{1}_{i,j}=1$ if $i$ and $j$ share a class label under the sampled attribute category and $0$ otherwise.  

This objective encourages each transformed embedding $\vv_i'$ to align closely with embeddings $\vv_j$ from the same attribute label within the base space. The query-only similarity score $\langle \mW\vv_i'+\vb, \vv\rangle$ factors as $\vv_i'^\top \mW^\top \vv + \vb^\top \vv$. The product $\vv_i'^\top \mW^\top \vv$ shows that the transformation is applied to the gallery side $\vv$ via the inner product at score time rather than rewritten into the stored gallery vectors. Thus, the gallery can remained fixed while the retrieval score is axis-aware.

\noindent\textbf{Multimodal Objective.} 
To maintain cross-modal consistency, we also include a symmetric multimodal alignment loss between transformed visual embeddings and text embeddings of the corresponding attribute labels:
\begin{equation}
\label{eq:multimodal}
\mathcal{L}_{i2t} = -\tfrac{1}{N} \sum_{i=1}^N 
\log \frac{\exp(\vv_i^{'\top} \bar{\vt}_i / \tau)}{\sum_{j=1}^N \exp(\vv_i^{'\top} \bar{\vt}_j / \tau)},
\end{equation}
where $\bar{\vt}_i$ represents an averaged text embedding over descriptive templates (e.g., ``a photo of \{c\}'', , ``an image in the \{c\} style''). $\mathcal{L}_{t2i}$ is defined in a similar fashion, just with $\vv'$ swapped with $\bar{\vt}$. 

The final loss combines the multi-positive visual term and the multimodal guidance terms:
\begin{equation}
    \label{eq:overall}
    \mathcal{L} = \mathcal{L}_{SR} + \omega*(\mathcal{L}_{i2t}+\mathcal{L}_{t2i}),
\end{equation}
where $\omega$ balances text alignment and category structure.  

\noindent\textbf{Dual Inference Capability.}
Although the hypernetwork is trained only in the \textit{query} setting, where transformed embeddings $\vv'$ are aligned to the frozen base space, it can be used for a dual purpose during inference time. The first straightforward application is to transform the query image, yielding an attribute-aware retrieval along the direction of the transformation. The second ability is to achieve an emergent \textit{global} transformation that reorganizes the entire distribution along the specified attribute axis without additional training. This transformed distribution shapes clustering to identify meaningful groupings based on the provided attribute. 
Concretely, the same learned transformation can be applied directly to the entire embedding set ($\mV \to \mV'$) at inference time. To summarize, the model learns a general operator that adapts to any visual embedding collection in an attribute-aware manner, enabling both attribute-aware retrieval and multi-attribute clustering within a single unified framework.

\begin{table}
    \centering
\begin{minipage}{0.49\linewidth}
    \centering
    \caption{Frozen gallery attribute-aware retrieval results reported as mAP averaged across dataset taxonomies. Our method simultaneously handles multiple categories with one set of weights while drastically improving performance. \dag~denotes MLP adapter trained with multi-positive loss.}
    \resizebox{\linewidth}{!}{
    \label{tab:main_retrieval}
\begin{tabular}{lcccc} 
\toprule
Method & \begin{tabular}[c]{@{}c@{}}Multi-Attr.\\Capability\end{tabular} & \begin{tabular}[c]{@{}c@{}}Clevr4\\mAP\end{tabular} & \begin{tabular}[c]{@{}c@{}}S40\\mAP\end{tabular} & \begin{tabular}[c]{@{}c@{}}ShotBench\\mAP\end{tabular} \\ 
\hline
Base Features & \xmark & 33.8 & 43.7 & 25.2 \\
LCR & \cmark & 15.9 & 19.9 & 18.9 \\
CWP & \cmark & 43.8 & 49.8 & 21.4 \\
ArcFace~\cite{deng2019arcface}\dag & \xmark & 12.3 & 18.8 & 17.5 \\
SupCon~\cite{khosla2020supervised}\dag & \xmark & 12.7 & 18.8 & 17.7 \\
StableRep+~\cite{tian2023stablerep}\dag & \xmark & 49.5 & 21.3 & 20.9 \\
MaPLe~\cite{khattakMaPLe} & \xmark & 33.5 & 47.8 & 23.7 \\
LoRA~\cite{hu2022lora} & \xmark & 42.9 & 49.3 & 22.0 \\
SEARLE-XL~\cite{baldrati2023zero} & \xmark & 24.6 & 33.9 & 20.3 \\
\textbf{Ours-Query} & \cmark & \textbf{77.0} & \textbf{86.3} & \textbf{41.0} \\
\bottomrule
\end{tabular}

    }
\end{minipage}
\hfill
\begin{minipage}{0.49\linewidth}
    \centering
    \caption{Multi-clustering results averaged over 14 attribute categories. Inference time is reported assuming a small evaluation set of 5k images. The proposed approach improves performance while offering minimal compute overhead and near-instant inference. (*reported on full action set.)}
    \resizebox{\linewidth}{!}{
    \label{tab:main_cluster}
\begin{tabular}{lcccc} 
\toprule
Method & \begin{tabular}[c]{@{}c@{}}Inference\\Time (s)\end{tabular} & \begin{tabular}[c]{@{}c@{}}Clevr4\\c. ACC\end{tabular} & \begin{tabular}[c]{@{}c@{}}S40\\c. ACC\end{tabular} & \begin{tabular}[c]{@{}c@{}}ShotBench\\c.ACC\end{tabular} \\ 
\hline
Base Features & 0.59 & 48.8 & 65.2 & 20.8 \\
LCR & 0.62 & 56.4 & 64.8 & 30.2 \\
CWP & 0.68 & 60.3 & 64.9 & 32.2 \\
Multi-MAP~\cite{yao2024multi} & 750 & 62.8 & 62.8 & 32.6 \\
Multi-Sub~\cite{yao2024customized} & 15300 & 72.2 & 66.1 & 32.4 \\
IC\textbar{}TC~\cite{kwon2023image} & 37680 & 57.9 & 76.1 & 21.7 \\
SSD-LLM~\cite{luo2025llm} & 27600 & 56.5 & 74.1 & 25.0 \\
X-Cluster~\cite{liu2025organizing} & 104760 & 64.9 & 68.3* & - \\
SEARLE-XL~\cite{baldrati2023zero} & 81 & 30.7 & 46.9 & 24.5 \\
\textbf{Ours-Space} & 0.64 & \textbf{73.0} & \textbf{80.0} & \textbf{51.6} \\
\bottomrule
\end{tabular}
    }
\end{minipage}
\end{table}

\section{Experiments}
\label{sec:experiments}

We first describe the experimental setup, then present results on attribute-aware retrieval, clustering, and show comprehensive ablation studies and analysis. Additional experimental results can be found in \supp{Supplementary Section C.}

\subsection{Datasets}
\noindent\textbf{ShotBench:} ShotQA/ShotBench~\cite{liu2025shotbench} are datasets designed for training and evaluating vision-language models on question-answering related to fine-grained photographic attributes, namely camera angle, composition, lens size, lighting, lighting type, shot framing, and shot size. From these, we extract attribute answers as labels, resulting in a multi-label dataset consisting of 31k training images and 2.7k testing images across 7 categories.

\noindent\textbf{Clevr-4~\cite{vaze2023clevr4}} is a synthetic dataset of 10k image samples, where each image contains a set of shapes with 4 axes of variation: color, count, shape, and texture. Each attribute category contains 10 subclasses and a mainly even distribution, so no subclass is over/underrepresented.

\noindent\textbf{Stanford 40 Actions~\cite{yao2011human}} contains 9,532 images of humans performing one of 40 actions, along with attribute annotations for 1,000 images with 10 location categories and 4 mood categories~\cite{kwon2023image}, resulting in three distinct categories: action, location, and mood. For training, we use a subset of 532 images drawn from the original action training split, annotating them with labels consistent with the existing taxonomy. Evaluation follows the same setup as prior work, using the full mood and location subsets and the remaining 9,000 images for action.

\noindent\textbf{MS-COCO~\cite{lin2014microsoft}.} We additionally use the COCO 2017 test split (41k images) for qualitative examples, serving as a diverse, real-world benchmark to illustrate cross-domain generalization of attribute transformations learned on synthetic or domain-specific datasets.

Additional information on datasets used may be found in \supp{Supp. Section A}.

\subsection{Baselines}
Few existing methods allow controllable adaptation of frozen embedding spaces, so we construct two lightweight, \textit{training-free} baselines to contextualize the benefits of our learned transformation. These baselines operate entirely using CLIP text and image embeddings and provide inexpensive, interpretable attribute–specific projections. 

Given an attribute category with $K$ class labels, we form a concept matrix
$\mA = [\vt_1, \ldots, \vt_K] \in \mathbb{R}^{K \times D}$,
where each row $\vt_k$ is the normalized text embedding of one attribute label. Both baselines project image embeddings $\mV \in \mathbb{R}^{N \times D}$ into the subspace spanned by these concept vectors.

\noindent\textbf{(1) Linear Concept Reconstruction (LCR).}
A simple reconstruction of each visual embedding in the concept span:
\begin{equation}
    \mV_{\text{LCR}} = (\mV \mA^T)\mA.
\end{equation}
LCR treats the attribute directions as a basis and re-expresses each image embedding as a weighted combination of them.

\noindent\textbf{(2) Concept-Weighted Projection (CWP).}
Because the concept vectors are not orthogonal, we correct for overlap via the inverse Gram matrix $(\mA \mA^T)^{-1}$:
\begin{equation}
    \mV_{\text{CWP}} = (\mV \mA^T)(\mA \mA^T)^{-1}\mA.
\end{equation}
This performs the closest-point projection onto the attribute subspace, effectively removing redundancy between overlapping concept directions.

Both baselines offer parameter-free, low-cost attribute awareness, but lack the learnable expressive ability of our proposed text-conditioned hypernetwork.

\subsection{Implementation Details}
All experiments are implemented in PyTorch~\cite{paszke2019pytorch} and trained on a single NVIDIA A100 GPU. The hypernetwork is a three-layer MLP with ReLU activations, which takes as input the CLIP text embedding and predicts the parameters of the affine transformation $(\mW, \vb)$. Dimensions depend on the chosen CLIP variant (e.g., 768 for ViT-L/14). We train with the AdamW~\cite{kingma2014adam, loshchilov2017decoupled} optimizer, a learning rate of $1\mathrm{e}{-4}$, and multimodal loss weight $\omega = 0.1$. All reported runs use frozen encoder weights; only the hypernetwork parameters are updated during training. Further implementation details are provided in \supp{Supp. Section B}.

\subsection{Attribute-aware Image-to-Image Retrieval}
\label{sec:retrieval}

We begin with the attribute-aware retrieval setting, where only the query embedding is transformed while retrieval is performed over the original frozen gallery. This \textit{query} mode preserves compatibility with large-scale search systems, as no re-encoding per attribute is required. Table~\ref{tab:main_retrieval} shows that CLIP’s base features do not offer control and struggle to surface attribute-consistent matches. Lightweight adapter-based approaches such as LoRA~\cite{hu2022lora} and MaPLe~\cite{khattakMaPLe} finetune CLIP to generally improve alignment. However, these require separate sets of parameters per attribute category, causing memory and storage costs to scale linearly with attribute count, and fail to handle the frozen gallery. In contrast, our method achieves state-of-the-art performance in this setting, while natively handling multiple categories and keeping memory footprint constant: a single hypernetwork handles all categories jointly, without growing in size as the number of attributes increases. Expanded results are provided in \supp{Supp. Section C}.

\subsection{Unsupervised Multi-Clustering}
\label{sec:clustering}

We next evaluate the \textit{global} setting, where the learned transformation is applied to the entire embedding set and unsupervised $k$-means clustering is performed on the transformed features. For each attribute, we run Faiss $k$-means~\cite{douze2024faiss} and compute clustering accuracy using Hungarian matching~\cite{kuhn1955hungarian}. Table~\ref{tab:main_cluster} compares our approach with state-of-the-art multi-clustering methods including Multi-MaP~\cite{yao2024multi}, Multi-Sub~\cite{yao2024customized}, IC$|$TC~\cite{kwon2023image}, SSD-LLM~\cite{luo2025llm}, and X-Cluster~\cite{liu2025organizing}. Methods that rely on image captioning and LLM iteration~\cite{kwon2023image,luo2025llm,liu2025organizing} incur high inference cost because each image must be captioned and repeatedly queried through a large language model, which are operations that scale linearly with dataset size and dominate runtime. \cite{yao2024customized,yao2024multi} reduce this reliance but still require per-attribute finetuning after user input, yielding moderate performance but still with substantial compute and memory overhead. In contrast, our proposed query-focused training method naturally handles this attribute disentanglement setting and achieves the highest average clustering accuracy across datasets while using a \textit{single} transformation network trained once for all attributes.

\begin{figure*}
    \centering
    \includegraphics[width=0.95\linewidth]{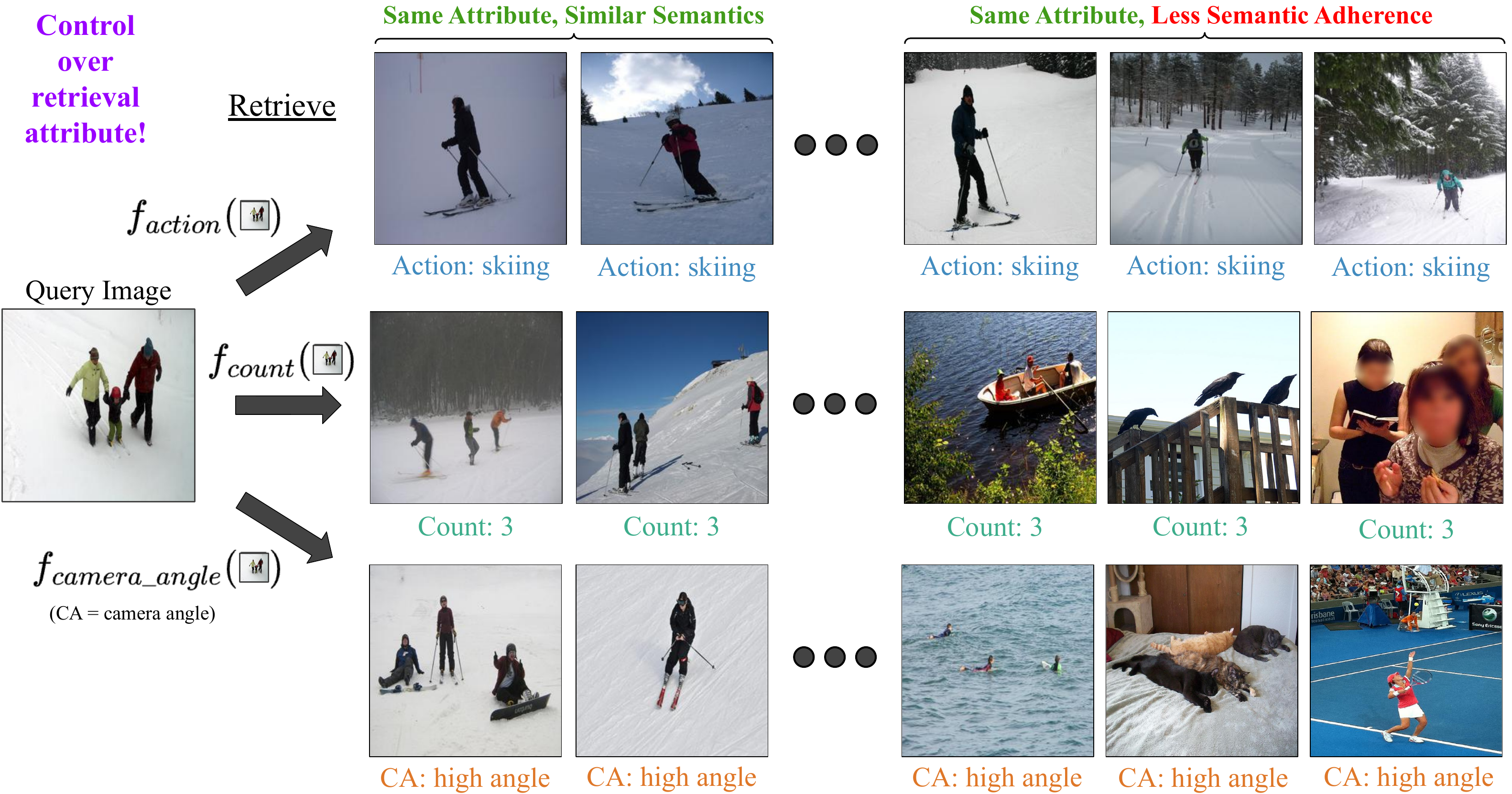}
    \vspace{-1.5mm}
    \caption{Qualitative attribute-aware retrieval results under different attribute transformation functions. Our text-conditioned query transformations selectively retrieve images that match the specified attribute without losing the scene semantics at first, then prioritize the attribute over general semantics. From top to bottom, we show retrievals for "action", "count", then fine-grained perceptual property "camera angle".}
    \label{fig:qual_coco}
\end{figure*}

\subsection{Ablations and Analysis}

\noindent\textbf{Transformation Type.} Attributes are entangled in CLIP space, yet they remain linearly separable~\cite{bhalla2024interpreting}. We demonstrate that controllable similarity is achievable with an asymmetric affine transformation applied to the query vector alone. A simple offset vector (translation) does not have the ability to scale and rotate sub-spaces, which is needed to satisfy our setting constraints (Table~\ref{tab:affine_justification}).

\vspace{-1mm}
\begin{table}[h]
\centering
\footnotesize
\setlength{\tabcolsep}{3.2pt}
\renewcommand{\arraystretch}{0.85}
\caption{Affine transforms effectively enable per-attribute similarity.}
\label{tab:affine_justification}
\vspace{-2mm}
\begin{tabular}{llccc}
\toprule
Dataset & Variant & Cluster ACC & Query mAP & Global mAP \\
\midrule
\multirow{3}{*}{Clevr-4}
& Base      & 32.2 & 33.8 & -- \\
& Translation-only & 32.2 & 35.4 & 34.5 \\
& \textbf{Affine}    & \textbf{75.4} {\scriptsize\textcolor{green!50!black}{(+43.2)}}
             & \textbf{77.6} {\scriptsize\textcolor{green!50!black}{(+42.2)}}
             & \textbf{85.1} {\scriptsize\textcolor{green!50!black}{(+50.6)}} \\
\midrule
\multirow{3}{*}{Stanford-40}
& Base      & 46.4 & 43.6 & -- \\
& Translation-only & 55.6 & 43.8 & 44.5 \\
& \textbf{Affine}    & \textbf{79.4} {\scriptsize\textcolor{green!50!black}{(+23.8)}}
             & \textbf{84.4} {\scriptsize\textcolor{green!50!black}{(+40.6)}}
             & \textbf{87.7} {\scriptsize\textcolor{green!50!black}{(+43.2)}} \\
\bottomrule
\end{tabular}
\end{table}
\vspace{-1mm}

\noindent\textbf{Compositional Retrieval.} Our framework directly supports compositional retrieval without retraining. We compose a multi-attribute query by linearly blending the projected queries into $q_{\mathrm{blend}}=\alpha \left(W_{\mathrm{a}} q + b_{\mathrm{a}}\right)+(1-\alpha)\left(W_{\mathrm{b}} q + b_{\mathrm{b}}\right)$, where $a$ and $b$ are attributes such as action and location. Table~\ref{tab:composition} shows that a composed query ($\alpha=0.5$) improves joint-label (Both) retrieval over the raw baseline and either single-axis projection, indicating that the learned projections can be combined for multi-attribute user queries.

\begin{table}[ht]
\centering
\footnotesize
\setlength{\tabcolsep}{2pt}
\renewcommand{\arraystretch}{0.85}
\caption{Compositional retrieval on Stanford40 action (A) + location (L). Uniform blending improves joint-label retrieval w/o retraining.}
\label{tab:composition}
\vspace{-2mm}
\begin{tabular}{lccccc}
\toprule
Variant & mAP(Both) & Both@5 & A-only@5 & L-only@5 & Neither@5 \\
\midrule
Base & 48.5 & 45.1 & 21.5 & 15.9 & \textbf{17.5} \\
Action only & 70.0 & 57.4 & \textbf{40.7} & 0.9 & 1.0 \\
Location only & 21.2 & 15.6 & 2.8 & \textbf{76.4} & 5.3 \\
Linear Blend ($\alpha=0.5$) & \textbf{80.9} & \textbf{64.8} & 30.2 & 3.6 & 1.4 \\
\bottomrule
\end{tabular}
\end{table}

\noindent\textbf{Embedding Space Analysis.}
We analyze how the learned transformation supports attribute-aware retrieval from a frozen gallery. For each attribute, we decompose $\mW$ with SVD and identify the top left-singular directions, which represent the axes most amplified by the transform. Even though gallery embeddings are unchanged, the query-gallery score is dominated by these amplified directions, while suppressed directions contribute little. We provide two findings: \\
\textbf{1) The transformed space is lower-dimensional and attribute-specific.} Stable rank, an effective-dimensionality measure, shows that projected features $\vv_i \mW^T$ occupy a smaller subspace than raw embeddings (Tab.~\ref{tab:stable_rank}). We further project features onto the dominant singular subspace of $\mW$ (retaining 95\% of singular-value energy) and its orthogonal complement. Retrieval in the kept subspace greatly exceeds the raw baseline, while retrieval in the orthogonal complement collapses, indicating that $\mW$ isolates a compact attribute-specific subspace.\\
\textbf{2) Query-only projection improves score discriminability.} Although absolute cosine similarities decrease after projection (Tab.~\ref{tab:class_sep}), same-label and different-label pairs become much better separated, and AUC improves along the target axis. This suggests that the transform suppresses irrelevant components rather than simply increasing all similarities.

\begin{table}[h]
\centering
\footnotesize
\renewcommand{\arraystretch}{0.85}

\begin{minipage}[t]{0.48\linewidth}
\centering
\caption{Stable rank and retrieval analysis. ``Kept subspace'' projects onto the dominant left-singular directions of $\mW$. ``Orth. Comp.'' indicates the orthogonal complement. The transform selects a compact attribute-specific subspace.}

\label{tab:stable_rank}
\begin{tabular}{lcccc}
\toprule
Axis & \makecell{Stable Rank\\(base$\to$trans)} & \makecell{Base\\mAP} & \makecell{Kept\\Space} & \makecell{Orth.\\Comp.} \\
\midrule
Color  & $11.3 \to 5.0$  & 11.9 & \textbf{98.4} & 4.9  \\
Action & $22.1 \to 14.3$ & 48.9 & \textbf{92.1} & 15.3 \\
\bottomrule
\end{tabular}
\end{minipage}
\hfill
\begin{minipage}[t]{0.50\linewidth}
\centering
\caption{Average cosine similarities for same/diff.-label pairs on Clevr4. AUC measures whether same-label pairs rank above different-label pairs. Transformed features drastically improve inter-class separation.}
\label{tab:class_sep}
\begin{tabular}{llccc}
\toprule
Axis & Mode & Same/Diff. & Cohen's \textit{d} & AUC \\
\midrule
\multirow{2}{*}{Color}
& Raw   & 0.76 / 0.72  & 0.7 & 0.703 \\
& Query & 0.19 / -0.02 & \textbf{4.0} & \textbf{0.993} \\
\midrule
\multirow{2}{*}{Count}
& Raw   & 0.74 / 0.72 & 0.3 & 0.579 \\
& Query & 0.10 / 0.01 & \textbf{1.4} & \textbf{0.829} \\
\bottomrule
\end{tabular}

\end{minipage}

\end{table}

\noindent\textbf{Generalization to Unseen Classes.} \Cref{tab:color_generalization} evaluates color-conditioned retrieval when specific color subclasses are held out during training. Our transformed embeddings still retrieve images of the same color as the query, despite not seeing it during training, indicating that the transformation successfully emphasizes the general color property instead of only learning the specific seen colors. However, similar colors (blue/cyan) struggle to fully disentangle.

\begin{table}[h]
\centering
\caption{Color generalization evaluation on Clevr4. Ours (seen) is the upper bound where all colors are present during training. Ours (unseen) indicates training with one color held out (e.g., no green), then evaluating color-based retrieval (mAP) on that color. Our transformation functions learn to generalize beyond the training subclasses.}
\label{tab:color_generalization}
\setlength{\tabcolsep}{1.35pt}
\begin{tabular}{lccccccccccc}
\toprule
                                                                                  & \multicolumn{11}{c}{Clevr4 – Color Retrieval (mAP)}                                                                                                                                                                                                                                                                                                                                                                                                                                                                                                                                                                                                                                                                                                                                                                                                   \\ 
\cmidrule(l){2-12}
Method                                                                            & Blue                                                                    & Brown                                                                    & Cyan                                                                    & Gray                                                                     & Green                                                                    & Orange                                                                   & Pink                                                                     & Purple                                                                   & Red                                                                      & Yellow                                                                   & Avg.                                                                      \\ 
\midrule
\rowcolor[rgb]{0.882,0.875,0.875} \textcolor[rgb]{0.502,0.502,0.502}{Ours (seen)} & \textcolor[rgb]{0.502,0.502,0.502}{99.6}                                & \textcolor[rgb]{0.502,0.502,0.502}{93.9}                                 & \textcolor[rgb]{0.502,0.502,0.502}{99.6}                                & \textcolor[rgb]{0.502,0.502,0.502}{99.3}                                 & \textcolor[rgb]{0.502,0.502,0.502}{99.6}                                 & \textcolor[rgb]{0.502,0.502,0.502}{86.2}                                 & \textcolor[rgb]{0.502,0.502,0.502}{88.5}                                 & \textcolor[rgb]{0.502,0.502,0.502}{99.6}                                 & \textcolor[rgb]{0.502,0.502,0.502}{98.4}                                 & \textcolor[rgb]{0.502,0.502,0.502}{74.2}                                 & \textbf{\textcolor[rgb]{0.502,0.502,0.502}{93.9}}                         \\
Baseline                                                                          & 23.3                                                                    & 16.9                                                                     & 26.5                                                                    & 21.0                                                                     & 28.6                                                                     & 21.0                                                                     & 20.4                                                                     & 30.1                                                                     & 21.5                                                                     & 19.4                                                                     & \textbf{22.9}                                                             \\
Ours (unseen)                                                                     & \makecell[c]{33.1\\[-1pt]\textcolor[rgb]{0,0.502,0}{\scriptsize{+9.8}}} & \makecell[c]{74.5\\[-1pt]\textcolor[rgb]{0,0.502,0}{\scriptsize{+57.6}}} & \makecell[c]{32.6\\[-1pt]\textcolor[rgb]{0,0.502,0}{\scriptsize{+6.1}}} & \makecell[c]{47.4\\[-1pt]\textcolor[rgb]{0,0.502,0}{\scriptsize{+26.4}}} & \makecell[c]{91.9\\[-1pt]\textcolor[rgb]{0,0.502,0}{\scriptsize{+63.3}}} & \makecell[c]{84.2\\[-1pt]\textcolor[rgb]{0,0.502,0}{\scriptsize{+63.2}}} & \makecell[c]{44.4\\[-1pt]\textcolor[rgb]{0,0.502,0}{\scriptsize{+24.0}}} & \makecell[c]{90.2\\[-1pt]\textcolor[rgb]{0,0.502,0}{\scriptsize{+60.1}}} & \makecell[c]{42.3\\[-1pt]\textcolor[rgb]{0,0.502,0}{\scriptsize{+20.8}}} & \makecell[c]{39.5\\[-1pt]\textcolor[rgb]{0,0.502,0}{\scriptsize{+20.1}}} & \makecell[c]{\textbf{58.0}\\[-1pt]\textcolor[rgb]{0,0.502,0}{\scriptsize{\textbf{+35.1}}}}  \\
\bottomrule
\end{tabular}
\end{table}

\noindent\textbf{Effect of Multimodal Guidance.}
Beyond improving retrieval and clustering, multimodal alignment also enhances text-similarity classification accuracy (Table~\ref{tab:abl_mm}), reflecting tighter correspondence between transformed embeddings and attribute text features. Even without explicit supervision, the transformation exhibits meaningful alignment due to CLIP’s pretrained cross-modal structure.

\begin{table}
\centering
\small
\caption{Ablation on multimodal guidance across three tasks using native metrics. The multimodal objective (Eq.~\ref{eq:multimodal}) enables text-similarity classification and modestly improves retrieval and clustering performance. Stanford40 abbreviated as S40 for brevity.}
\label{tab:abl_mm}
\setlength{\tabcolsep}{0.8pt}
\begin{tabular}{lccccccccc}
\toprule
& \multicolumn{3}{c}{Multiple Clustering} 
& \multicolumn{3}{c}{Attr.-Aware Retrieval} 
& \multicolumn{3}{c}{Text-based Classification} \\
\cmidrule(lr){2-4} \cmidrule(lr){5-7} \cmidrule(lr){8-10}
Objective 
& Clevr4 & S40 & ShotBench 
& Clevr4 & S40 & ShotBench 
& Clevr4 & S40 & ShotBench \\
\midrule
\textbf{Ours}  
& \textbf{73.0} & \textbf{80.0} & 51.6
& \textbf{77.1} & \textbf{91.2} & \textbf{41.1}
& \textbf{88.7} & \textbf{83.2} & \textbf{51.6} \\
$-$ multimodal 
& 72.7 & 78.4 & \textbf{53.2}
& 76.8 & 87.3 & 40.7
& 71.7 & 78.5 & 31.3 \\
\bottomrule
\end{tabular}
\end{table}

\begin{figure}[ht]
    \centering

    \begin{minipage}[ht]{0.45\linewidth}
        \centering
        \includegraphics[width=\linewidth]{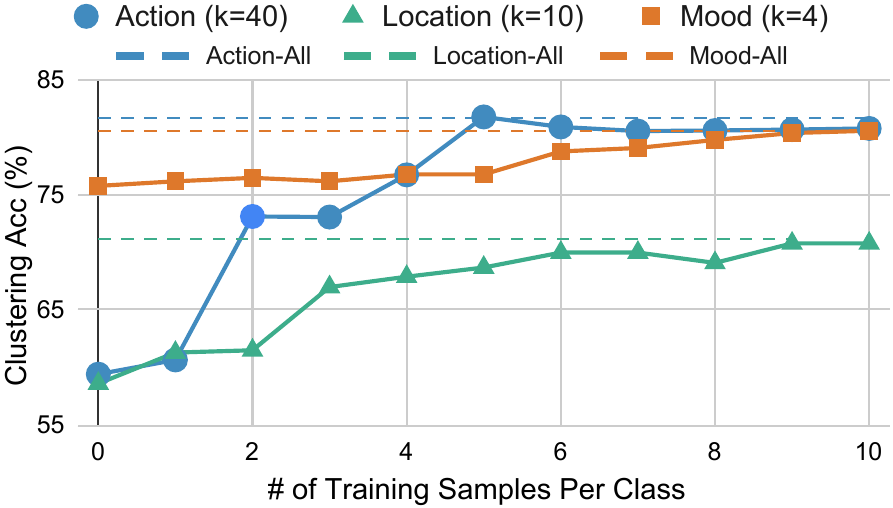}
        \caption{Multi-clustering accuracy on Stanford40 under varying number of training images per class. Category-All line indicates performance using all available training samples. Our transform approach demonstrates extreme data efficiency, achieving near peak results with just \textbf{six} images per class.}
        \label{fig:max_per_class}
    \end{minipage}\hfill
    \begin{minipage}[ht]{0.52\linewidth}
        \centering
        \vspace{4mm}
        \includegraphics[width=\linewidth]{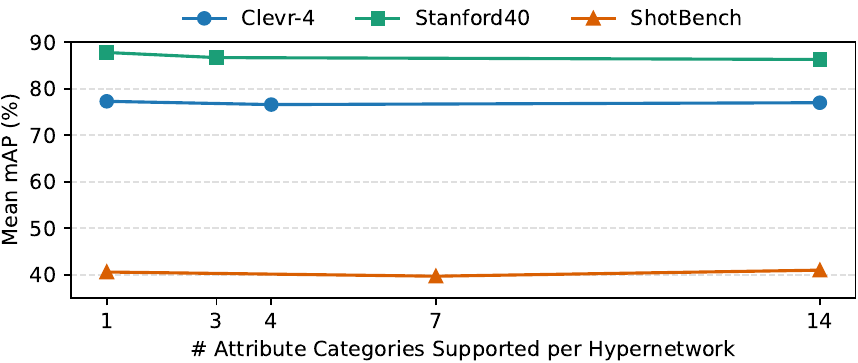}
        \caption{Mean category mAP across datasets (each line) as the number of jointly trained attribute categories increases from a single category per model to unified training across all datasets. Performance remains consistent even as more attributes are added, demonstrating that a single hypernetwork can scale across 14+ attributes without degradation.}
        \label{fig:abl_capacity}
    \end{minipage}
\end{figure}

\noindent\textbf{Data Efficiency.} Figure~\ref{fig:max_per_class} ablates the number of training samples used per class on the Stanford 40 Action~\cite{yao2011human} dataset. Remarkably, our global affine transformation achieves near peak performance across all taxonomies given just \textbf{six} training samples per class.

\noindent\textbf{Scalability Across Attributes.} 
We assess how well the model scales as more attribute categories are trained within a single hypernetwork. Starting from one model per attribute, we progressively expand to one per dataset and finally to a unified model trained across all datasets. The results in Figure~\ref{fig:abl_capacity} show that performance remains stable even as the number of jointly trained attributes increases, varying by $\leq$1.5\%. Notably, final results are presented using a single hypernetwork trained across \textbf{19} attributes with \textbf{187} subclasses, indicating that the hypernetwork can generalize across many attributes without interference.

\begin{figure}
    \centering
    \includegraphics[width=0.97\linewidth]{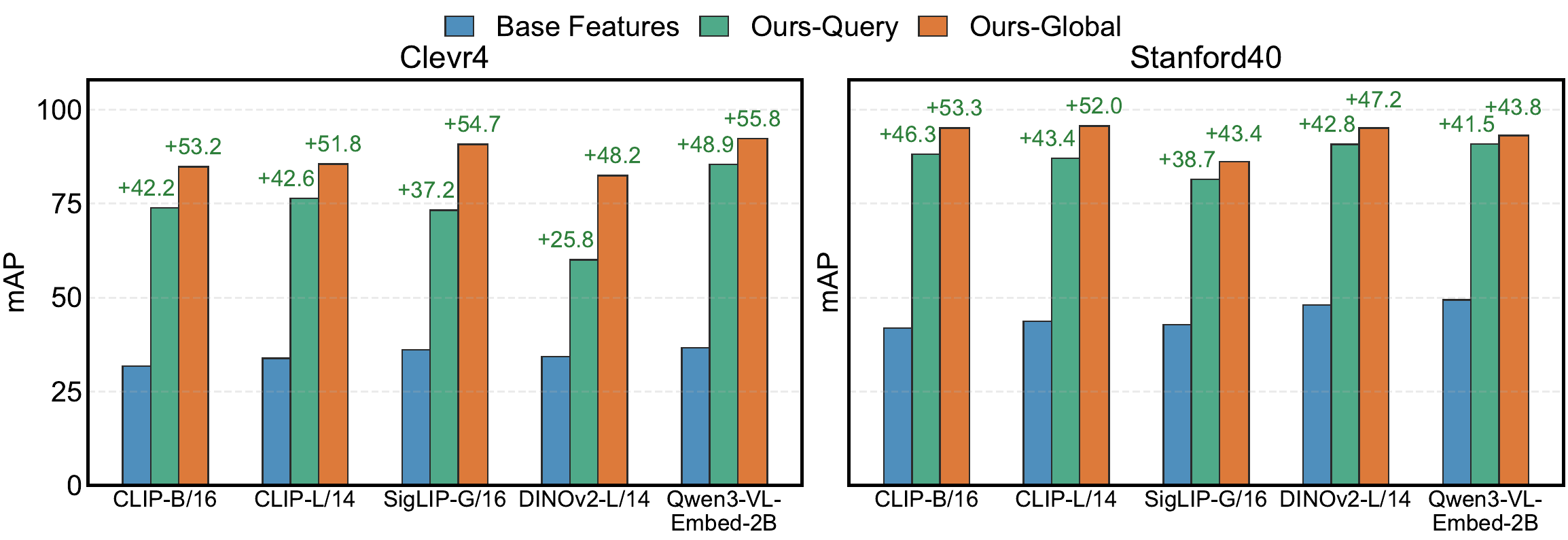}
    \caption{Ablation on backbone architecture, evaluated using retrieval mAP. Our method is effective across all types/sizes of encoders, including non-text aligned DINOv2.}
    \label{fig:abl_backbone}
\end{figure}

\noindent\textbf{Backbone Encoder Choice.}
Figure~\ref{fig:abl_backbone} shows that the proposed transformation consistently improves attribute-based retrieval across encoder architectures and model sizes. Gains are observed for both vision–language models (CLIP, SigLIP, Qwen3-VL-Embed) and purely visual backbones (DINOv2), demonstrating that the method generalizes beyond text-aligned representations. Note that for DINOv2, the CLIP H/14 text encoder is used for consistent feature dimensionality.

\subsection{Discussion}

We next analyze qualitative and emergent properties of learned transformations.

\noindent\textbf{How can a single transformed query embedding retrieve from a frozen gallery set?}
The transformation is trained with a multi-positive contrastive loss which optimizes relative similarity—pulling the transformed query closer to positives while repelling negatives. Because most concepts are partially entangled (e.g., two images may share an object but differ in color), minimizing absolute distance to positives would inadvertently bring it near some negatives. To resolve this, the network learns to move the query in a new direction within the high-dimensional space that separates it from the general visual manifold while aligning it with the target attribute. This broadly resembles the text–image “modality gap”~\cite{liang2022mind}, allowing the transformed query to behave similar to a text embedding. Even without explicit multimodal supervision, Table~\ref{tab:abl_mm} shows that transformed queries naturally align with text features, enabling accurate retrieval from a frozen gallery using similarity in the new space.

\begin{figure}
    \centering
    \includegraphics[width=0.8\linewidth]{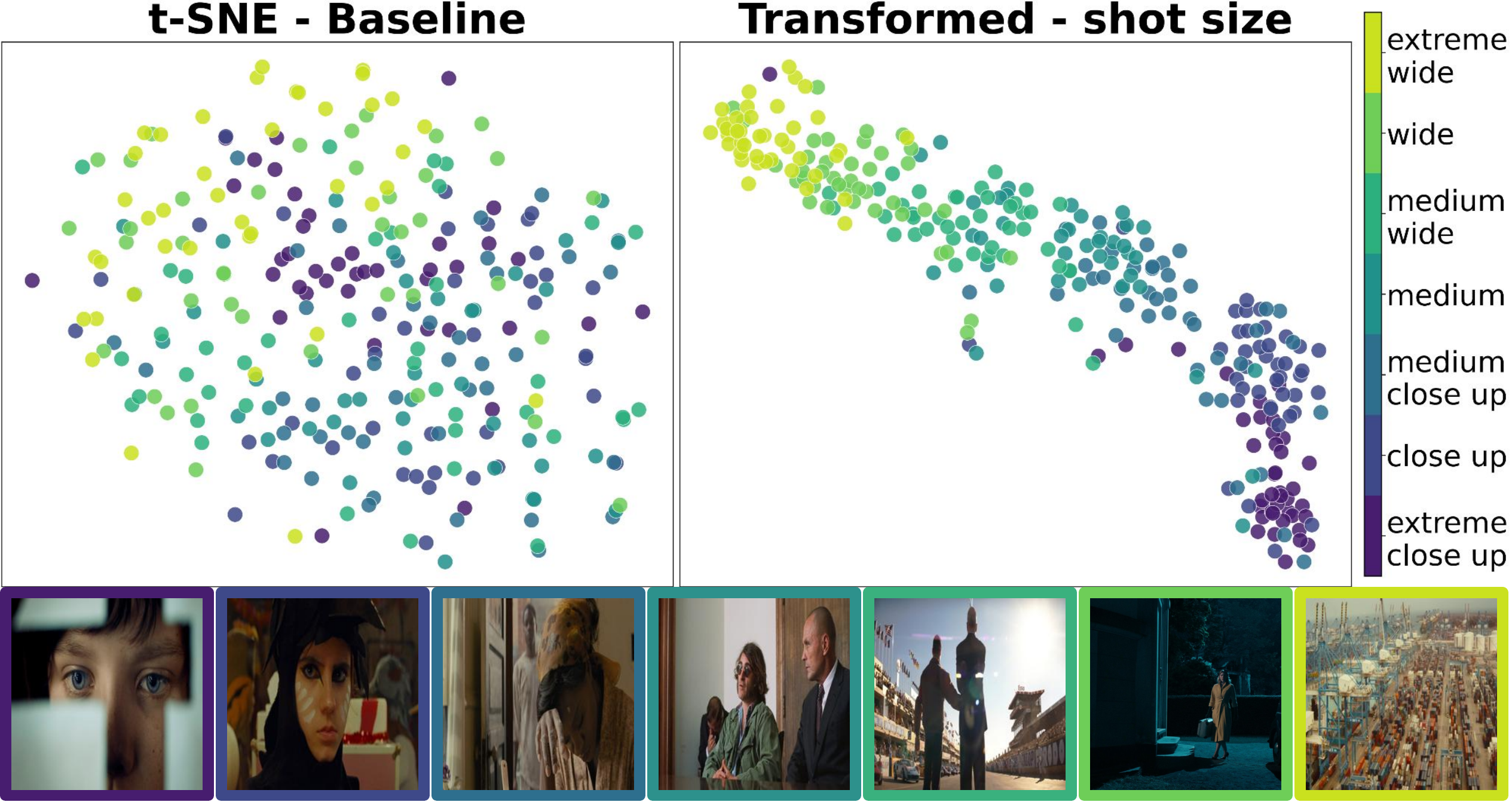}
    \caption{t-SNE visualization of embeddings colored by shot-size labels before and after transformation. The baseline space (left) shows entangled clusters. After applying our learned transformation (right), samples align intuitively along a continuous gradient that reflects the shot-size taxonomy.}
    \label{fig:tsne}
\end{figure}

\noindent\textbf{Can the transformation capture perceptual attributes and how does it affect embedding geometry?}
Contrastive image–text models such as CLIP are trained to capture broad semantic meaning, yet they provide little explicit separation for perceptual attributes like shot size, composition, or camera angle. As a result, embeddings that differ in these subtle properties often remain close in the feature space. Our transformation isolates these attributes by reorienting the space toward the specified taxonomy while preserving global structure. As shown in Figure~\ref{fig:tsne}, applying the transformation introduces a clear, continuous axis of variation corresponding to the target attribute (here, shot size). Samples form a smooth progression from close-up to wide shots, reflecting a structured and interpretable organization. Quantitatively, this behavior yields strong improvements on fine-grained perceptual benchmarks like ShotBench, confirming that the transformation reshapes latent geometry to make attribute-specific relationships explicit without distorting the original feature layout.

\begin{figure}
    \centering
    \includegraphics[width=\linewidth]{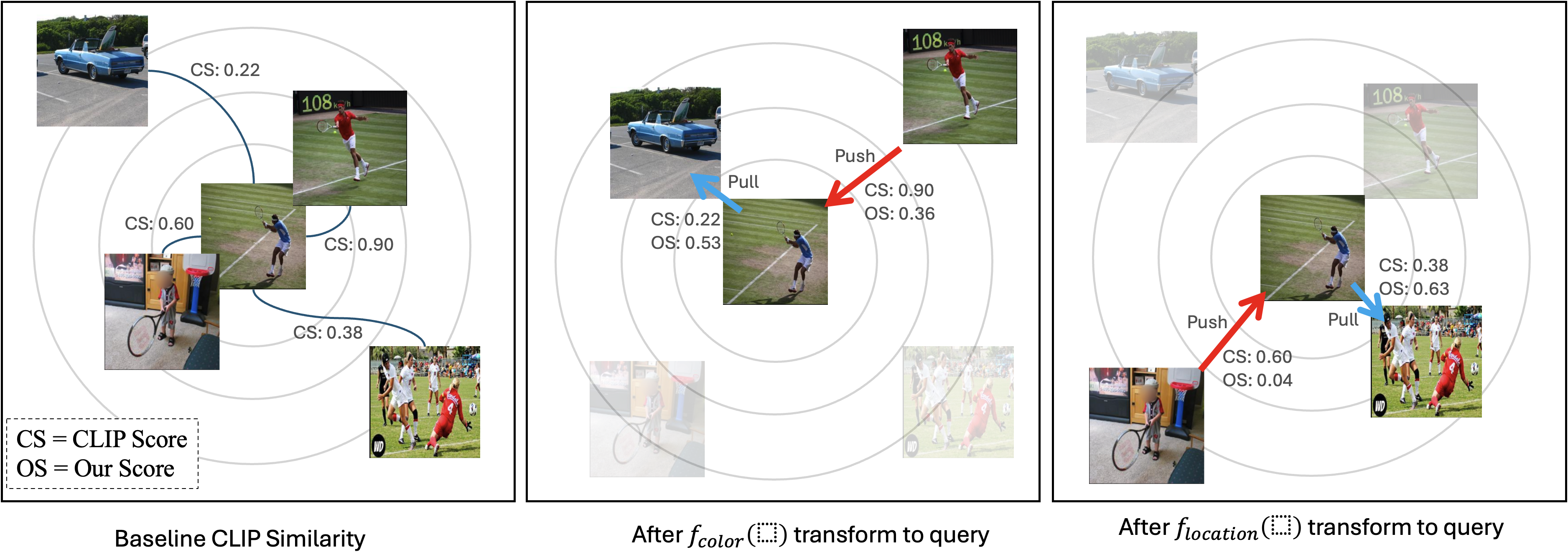}
    \caption{Our attribute-conditioned similarity metric helps characterize unique relationships that CLIP similarity alone cannot. (Left) Baseline CLIP similarity prioritizes overall semantics. (Middle, Right) Applying attribute-specific transforms to the query (e.g., color or location) selectively re-scores relationships based on the input attribute.}
    \label{fig:coco_sim_qual}
\end{figure}

\noindent\textbf{Do attribute-specific transforms learned on synthetic data transfer to real images?}
Our transformations generalize cleanly across domains. When trained on the synthetic Clevr-4 dataset, which contains simple geometric objects with uniform backgrounds, the model still produces meaningful attribute disentanglement for real, diverse MS-COCO images (Figs.~\ref{fig:qual_coco} and \ref{fig:coco_sim_qual}). Despite the distribution shift, transforms learned for color and count correctly align COCO embeddings along the intended dimensions. We conjecture that this generalization arises from optimizing a single affine transform for the entire embedding space, which discourages shortcut memorization and instead forces the transform to properly disentangle embeddings to optimize the contrastive loss.

\noindent\textbf{Attribute-aware similarity.} Figure~\ref{fig:coco_sim_qual} illustrates how the proposed similarity metric handles relationships that are not captured by standard CLIP similarity. Images that share the attribute are pulled closer while others are pushed away, irrespective of their original CLIP similarity. Our transform effectively restructures the embedding space to emphasize the selected attribute, enabling controlled comparison of images along different attribute dimensions.

\noindent\textbf{Composed Image Retrieval (CIR) Compatibility.}
Our transformations can also be integrated into existing CIR pipelines to improve attribute discriminability. We provide an illustrative experiment in \supp{Supp. Section C}.

\section{Conclusion}
\label{sec:conclusion}

We presented a text-conditioned transformation framework that enables fine-grained control over pretrained embedding spaces without modifying encoder weights or recomputing gallery features. A single hypernetwork learns to generate attribute-specific affine transformations from natural-language descriptions, supporting both attribute-aware retrieval and multi-attribute organization with near-zero inference overhead. While the method generalizes across diverse attributes, its expressiveness is currently bounded by the range of categories seen during training. Future work may explore scaling up training to find emergent transformation properties across attributes completely unseen during training.


%
%
\bibliographystyle{splncs04}
\bibliography{main}

\clearpage
\appendix
\enableindentfirst

\setcounter{table}{0}
\renewcommand{\thetable}{S\arabic{table}}

\setcounter{figure}{0}
\renewcommand{\thefigure}{S\arabic{figure}}

\section*{Supplementary Overview}

\Cref{sec:dataset_details}: Dataset details

\Cref{sec:imp_details}: Implementation details

\Cref{sec:exp_details}: Additional experiment details

\section{Dataset Details}
\label{sec:dataset_details}

\noindent\textbf{ShotBench~\cite{liu2025shotbench}.} ShotBench provides annotations for perceptual, camera-related attributes such as camera angle, shot size, composition, and lighting. Although the dataset was originally introduced for visual question answering, we convert the answer strings into categorical labels for each attribute. A small number of answers contain multiple labels for the same category (e.g., “low angle \& dutch angle”); these ambiguous cases are removed to ensure a single ground-truth label per category. We use the official split of 31k training images and 2.7k test images.

\noindent\textbf{Clevr-4~\cite{vaze2023clevr4}.} Clevr-4 is a synthetic benchmark constructed to isolate four independent visual attributes: color, count, shape, and texture. Each rendered image depicts multiple 3D objects, and each attribute category contains 10 uniformly sampled subclasses, giving the dataset a clean, balanced structure ideal for learning controlled transformations. We use the official split of 8.4k training images and 2.1k test images.

\noindent\textbf{Stanford 40 Actions~\cite{yao2011human}.} Stanford40 consists of 9,532 images spanning 40 human actions. Following prior multi-attribute clustering works \cite{kwon2023image}, we also use the additional annotations provided for two auxiliary categories: location (10 classes) and mood (4 classes), each defined for a disjoint subset of 1,000 images. To supply supervision for all three categories during training, we annotate a small set of 532 images from the original Stanford40 training split with labels consistent with the released location and mood definitions. These annotations do not overlap with the evaluation subsets. For evaluation, we follow the same protocol as prior work: action is evaluated on the remaining ~9,000 images, location and mood are evaluated on their respective 1,000 image annotated subsets.

\noindent\textbf{FruitSALAD~\cite{ohm2025FruitSALAD}.} FruitSALAD is a style–aligned artwork dataset designed to disentangle object identity from artistic appearance. It contains 10 fruit categories and 10 artistic styles, with each fruit–style pairing instantiated by 100 images (for a total of 10,000 samples). This structure makes it well suited for evaluating attribute-aware retrieval. We train with even splits, taking 50 of each pairing instance as training images and the remaining as test images.

\noindent\textbf{FER-2013~\cite{goodfellow2013challenges}.} FER-2013 is a facial expression recognition dataset containing 35,887 grayscale face images annotated with 7 emotion classes (anger, disgust, fear, happiness, sadness, surprise, neutral). We use the standard split: 28,709 training images and 3,589 public test images.

\noindent\textbf{Card~\cite{yao2023augdmc}.} Card is a playing-card recognition dataset with 8,029 images, annotated along two independent categories: suit (clubs, spades, diamonds, hearts) and rank (2–10, jack, queen, king, ace). We follow the official split of 7,624 training images, 265 validation images, and 265 test images.

\noindent\textbf{MS-COCO~\cite{lin2014microsoft}.} We additionally evaluate on images from the MS-COCO 2017 \textit{test2017} split (41k images) to study cross-domain generalization. Since COCO does not provide annotations for the attribute taxonomies used in this work, we generate labels using Qwen3-VL-32B with a structured prompt aligned to the attribute subclasses defined by the source datasets. The hypernetwork is trained only on CLEVR-4, Stanford40, and ShotBench, and applied to COCO without further training. This setting therefore evaluates zero-shot transfer of the learned attribute transformations to a diverse real-world image distribution.

\noindent\textbf{FashionIQ~\cite{wu2021fashion}.} FashionIQ is a benchmark for composed fashion image retrieval, where each query is defined by a reference image and a natural-language change text describing how to modify it to obtain a target image. The dataset is organized into three product categories (Dresses, Shirts, Tops\&Tees). We report results on the official validation split of 6,016 reference–text–target queries.

\noindent\textbf{SST-5~\cite{socher2013recursive}.} The Stanford Sentiment Treebank (SST) is a sentence-level sentiment dataset derived from movie reviews, where each sentence is annotated with one of five ordered sentiment classes ranging from very negative to very positive. We use the standard SST-5 splits of 8,544 training and 2,210 test sentences, extracting text embeddings for each sentence and learning sentiment-conditioned transformations in the same manner as our visual-attribute experiments.

\section{Implementation Details}
\label{sec:imp_details}

\noindent\textbf{Memory Footprint Estimation.} Table~\ref{tab:global_ret} includes a memory footprint estimation for maintaining embeddings across 14 distinct attriubtes. The raw storage cost for 1M ($N=10^6$) vectors, each of dimensionality $D=768$ and stored with 32-bit floating-point precision ($4$ bytes), is $N \times D \times 4 \text{ bytes} = 1,000,000 \times 768 \times 4 \approx 3.07 \approx 3 \text{ GB}$. Additionally, we assume an HNSW (Hierarchical Navigable Small World)~\cite{malkov2018efficient} index for fast search is used, incurring a memory overhead estimated at a factor of $2\times$ over the raw vector storage. This results in a total estimated in-memory footprint of $3 \text{ GB} \times 1.5 \approx 6 \text{ GB}$ for each 1M embeddings.

\section{Additional Experiments}
\label{sec:exp_details}

\noindent\textbf{Composed Image Retrieval (CIR) Compatibility.} To demonstrate that our query-only transformations can \emph{enhance} existing CIR pipelines, we augment SEARLE-XL~\cite{baldrati2023zero} on FashionIQ~\cite{wu2021fashion}. We keep the gallery fixed by precomputing normalized image embeddings $\vv_j = f_v(I_j)$ for all candidates. Given a CIR query (reference image $I_r$ and relative caption $T$), SEARLE-XL produces a composed query embedding $\vq \in \mathbb{R}^D$ in the same CLIP space. Baseline retrieval ranks images by $s_{\text{base}}(j)=\vq^\top \vv_j$. We then apply our \emph{query-only} transformation using the color attribute text to obtain $\vq'$, and define a color-aware auxiliary score $s_{\text{color}}(j)={\vq'}^\top \vv_j$. A lightweight intent parser (Llama~3.2~3B-Instruct) maps the caption $T$ to a signed weight $w\in[-1,1]$ indicating whether color should be preserved ($w>0$) or changed ($w<0$). Final ranking uses
\begin{equation}
s(j) \;=\; \alpha*s_{\text{base}}(j) \;+\; (1-\alpha)*\, w\, * s_{\text{color}}(j),
\end{equation}
with $\alpha=0.8$ and no training on FashionIQ. As shown in Table~\ref{tab:cir}, this simple plug-in reranking consistently improves SEARLE-XL across all categories, supporting that our attribute-consistency metric is compatible with CIR and enhances performance by injecting an explicit attribute prior.

\begin{table}
\centering
\caption{Composed image retrieval results on FashionIQ. \textbf{+Ours} denotes augmenting SEARLE-XL~\cite{baldrati2023zero} with our attribute-consistency metric scoring for just "color", leaving the gallery index unchanged and using no FashionIQ training.}
\label{tab:cir}
\begin{tabular}{l|cc|cc|cc|cc} 
\toprule
                     & \multicolumn{2}{c|}{Dress}    & \multicolumn{2}{c|}{Shirt}    & \multicolumn{2}{c|}{Toptee}   & \multicolumn{2}{c}{Average}    \\
Method               & R@10          & R@50          & R@10          & R@50          & R@10          & R@50          & R@10          & R@50           \\ 
\hline
SEARLE-XL~\cite{baldrati2023zero}            & 25.9          & 42.9          & 18.2          & 37.2          & 26.5          & 45.1          & 23.5          & 41.8           \\
~~~~ + Ours & \textbf{26.3} & \textbf{44.9} & \textbf{19.8} & \textbf{39.0} & \textbf{27.8} & \textbf{46.1} & \textbf{24.6} & \textbf{43.3}  \\
\bottomrule
\end{tabular}
\end{table}

\noindent\textbf{Text-based Attribute-aware Retrieval Results.} \Cref{tab:text_sentiment} shows that our embedding-space manipulation framework is not restricted to visual attributes. When applied to text embeddings on SST-5, learning a sentiment-conditioned transformation consistently improves performance across all evaluation modes. These gains indicate that the proposed mechanism generalizes across embedding spaces independent of the modality.

\begin{table}
\centering
\caption{Sentiment categorization on SST-5 using text embeddings (E5-large-v2). Our embedding-space transformation improves performance in all three attribute-aware tasks, demonstrating that our approach generalizes beyond vision to textual attributes.}
\label{tab:text_sentiment}
\begin{tabular}{lccc} 
\toprule
         & \multicolumn{3}{c}{SST5 - Sentiment}  \\
Method   & Retrieval & Clustering & Zero-Shot    \\ 
\hline
Baseline & 30.9      & 41.7       & 29.6         \\
Ours     & \textbf{44.5}      & \textbf{51.3}       & \textbf{54.3}         \\
\bottomrule
\end{tabular}
\end{table}

\noindent\textbf{Attribute Abbreviations.}  
For readability in tables, attribute categories are abbreviated as follows:

\begin{itemize}
    \item Clevr-4: Col = color, Cnt = count, Shp = shape, Tex = texture.
    \item Stanford 40: Act = action, Loc = location, Md = mood.
    \item ShotBench: CA = camera angle, CM = composition, LS = lens size, LI = lighting, LT = lighting type, SF = shot framing, SS = shot size.
\end{itemize}

\begin{table*}
\centering
\setlength{\tabcolsep}{2.5pt}
\caption{Category-wise attribute-aware retrieval mAP across multiple datasets. Category names are abbreviated (full names in Section~\ref{sec:exp_details}). Highlighted columns denote dataset averages. Our approach retrieves meaningful samples while transforming just the query embedding.}
\label{tab:retrieval_expanded}
\resizebox{\linewidth}{!}{
\begin{tabular}{lccccc|cccc|cccccccc} 
\toprule
 & \multicolumn{5}{c|}{Clevr-4} & \multicolumn{4}{c|}{Stanford 40 Action} & \multicolumn{8}{c}{ShotBench} \\
Method & Col & Cnt & Shp & Tex & Avg. & Act & Loc & Md & Avg. & CA & CM & LS & LI & LT & SF & SS & Avg. \\ 
\midrule
Base Features & 23.0 & 13.9 & 79.5 & 19.0 & {\cellcolor[rgb]{0.855,0.914,0.973}}33.8 & 49.8 & 40.2 & 41.2 & {\cellcolor[rgb]{0.855,0.914,0.973}}43.7 & \underline{34.5} & \underline{21.2} & \underline{32.4} & \underline{18.2} & \underline{22.5} & \underline{24.9} & 22.7 & {\cellcolor[rgb]{0.855,0.914,0.973}}\underline{25.2} \\
LCR & 18.4 & 11.3 & 19.2 & 14.8 & {\cellcolor[rgb]{0.855,0.914,0.973}}15.9 & 6.3 & 24.8 & 28.5 & {\cellcolor[rgb]{0.855,0.914,0.973}}19.9 & 22.5 & 17.9 & 29.6 & 15.0 & 11.9 & 16.2 & 19.0 & {\cellcolor[rgb]{0.855,0.914,0.973}}18.9 \\
CWP & 67.4 & 14.2 & 73.1 & 20.4 & {\cellcolor[rgb]{0.855,0.914,0.973}}43.8 & \underline{72.2} & 44.9 & 32.3 & {\cellcolor[rgb]{0.855,0.914,0.973}}49.8 & 24.6 & 19.4 & 29.6 & 17.0 & 15.2 & 21.8 & 22.6 & {\cellcolor[rgb]{0.855,0.914,0.973}}21.4 \\
ArcFace\cite{deng2019arcface}\dag & 10.9 & 10.3 & 17.5 & 10.7 & {\cellcolor[rgb]{0.855,0.914,0.973}}12.3 & 3.7 & 24.0 & 28.8 & {\cellcolor[rgb]{0.855,0.914,0.973}}18.8 & 21.7 & 17.6 & 26.3 & 13.6 & 11.9 & 16.0 & 15.6 & {\cellcolor[rgb]{0.855,0.914,0.973}}17.5 \\
SupCon\cite{khosla2020supervised}\dag & 13.2 & 13.0 & 14.5 & 10.2 & {\cellcolor[rgb]{0.855,0.914,0.973}}12.7 & 3.2 & 24.5 & 28.7 & {\cellcolor[rgb]{0.855,0.914,0.973}}18.8 & 22.8 & 17.8 & 27.4 & 13.8 & 11.4 & 15.0 & 15.8 & {\cellcolor[rgb]{0.855,0.914,0.973}}17.7 \\
StableRep+\cite{tian2023stablerep}\dag & \underline{79.6} & \underline{27.4} & 62.2 & 28.6 & {\cellcolor[rgb]{0.855,0.914,0.973}}\underline{49.5} & 10.2 & 24.7 & 29.2 & {\cellcolor[rgb]{0.855,0.914,0.973}}21.3 & 23.4 & 17.6 & 29.6 & 14.8 & 14.2 & 23.7 & 23.1 & {\cellcolor[rgb]{0.855,0.914,0.973}}20.9 \\
MaPLe\cite{khattakMaPLe} & 23.9 & 16.2 & 70.2 & 23.6 & {\cellcolor[rgb]{0.855,0.914,0.973}}33.5 & 45.6 & 44.7 & 53.3 & {\cellcolor[rgb]{0.855,0.914,0.973}}47.8 & 32.5 & 20.3 & 30.6 & 17.4 & 20.8 & 23.0 & 21.4 & {\cellcolor[rgb]{0.855,0.914,0.973}}23.7 \\
LoRA\cite{hu2022lora} & 34.4 & 17.7 & \underline{83.4} & \underline{36.0} & {\cellcolor[rgb]{0.855,0.914,0.973}}42.9 & 46.5 & \underline{46.2} & \underline{55.3} & {\cellcolor[rgb]{0.855,0.914,0.973}}\underline{49.3} & 30.1 & 17.9 & 31.1 & 14.7 & 12.4 & 24.1 & \underline{23.8} & {\cellcolor[rgb]{0.855,0.914,0.973}}22.0 \\
\textbf{Ours-Query} & \textbf{93.4} & \textbf{39.3} & \textbf{99.4} & \textbf{75.9} & {\cellcolor[rgb]{0.855,0.914,0.973}}\textbf{77.0} & \textbf{89.3} & \textbf{83.9} & \textbf{85.7} & {\cellcolor[rgb]{0.855,0.914,0.973}}\textbf{86.3} & \textbf{52.}7 & \textbf{28.0} & \textbf{48.4} & \textbf{22.8} & \textbf{31.8} & \textbf{57.8} & \textbf{45.8} & {\cellcolor[rgb]{0.855,0.914,0.973}}\textbf{41.0} \\
\bottomrule
\end{tabular}
}
\end{table*}
\noindent\textbf{Expanded Attribute-aware Retrieval Results.}
To supplement the dataset-level retrieval results in the main paper, we also report mAP for each attribute category in Table~\ref{tab:retrieval_expanded}. These per-attribute measurements reveal how retrieval performance varies across attributes and demonstrate that the query transformation consistently improves attribute alignment across a broad range of properties.

\begin{table}[htb]
\centering
\caption{Mean average precision (mAP) retrieval performance across other benchmarks. Top section is the query (frozen gallery) setting, bottom section is the global transformation setting. Scores are averaged across taxonomies within each dataset. Our method strongly improves retrieval across the board.}
\label{tab:extra_mAP}
\setlength{\tabcolsep}{3.0pt}
\small
\begin{tabular}{lcc|c|cc|c} 
\toprule
\multirow{2}{*}{Method} & \multicolumn{2}{c|}{Cards} & FER & \multicolumn{2}{c|}{Fruits} & \multirow{2}{*}{\begin{tabular}[c]{@{}c@{}}All\\Avg.\end{tabular}} \\
 & \multicolumn{1}{l}{Rank} & \multicolumn{1}{l|}{Suit} & \multicolumn{1}{l|}{Emotion} & \multicolumn{1}{l}{Fruit} & \multicolumn{1}{l|}{Style} &  \\ 
\midrule
Base Features & 37.9 & 44.6 & 25.6 & 67.7 & 58.8 & {\cellcolor[rgb]{0.855,0.914,0.973}}43.4 \\
LCR & 10.4 & 31.0 & 18.1 & 28.1 & 23.1 & {\cellcolor[rgb]{0.855,0.914,0.973}}21.5 \\
CWP & 55.5 & 48.0 & 18.5 & 97.6 & 53.3 & {\cellcolor[rgb]{0.855,0.914,0.973}}48.6 \\
\textbf{Ours-Query} & \textbf{86.0} & \textbf{96.2} & \textbf{52.7} & \textbf{99.8} & \textbf{99.7} & {\cellcolor[rgb]{0.855,0.914,0.973}}\textbf{81.2} \\ 
\midrule
Base Features & 37.9 & 44.6 & 25.6 & 67.7 & 58.8 & {\cellcolor[rgb]{0.855,0.914,0.973}}43.4 \\
LCR & 54.6 & 85.4 & 18.7 & 96.8 & 83.4 & {\cellcolor[rgb]{0.855,0.914,0.973}}59.6 \\
CWP & 80.2 & 84.9 & 19.1 & 99.1 & 84.5 & {\cellcolor[rgb]{0.855,0.914,0.973}}64.5 \\
\textbf{Ours-Global} & \textbf{90.9} & \textbf{97.9} & \textbf{56.0} & \textbf{99.8} & \textbf{99.8} & {\cellcolor[rgb]{0.855,0.914,0.973}}\textbf{83.4} \\
\bottomrule
\end{tabular}
\end{table}
\noindent\textbf{Additional Datasets: Attribute-aware Retrieval.}
We report results on three additional benchmarks (Card, FER, and FruitSALAD) to demonstrate broad attribute coverage in Table~\ref{tab:extra_mAP}. Across both query-only mode and full space inference, our method maintains strong mAP performance.

\begin{table}
\centering
\setlength{\tabcolsep}{2.5pt}
\caption{Recall@1 and Recall@5 attribute-aware retrieval performance using a frozen gallery. Results are averaged over all categories within each dataset. Our method consistently surfaces an attribute-matching image within the top retrieved results.}
\label{tab:recall}
\begin{tabular}{lcl|cl|cl} 
\toprule
 & \multicolumn{2}{c|}{Clevr4} & \multicolumn{2}{c|}{Stanford40} & \multicolumn{2}{c}{ShotBench} \\
Method & R@1 & R@5 & R@1 & R@5 & R@1 & R@5 \\ 
\hline
Base Features & 70.1 & 88.6 & 72.7 & 92.4 & 40.8 & 77.2 \\
LCR & 19.2 & 34.6 & 9.8 & 48.4 & 17.5 & 47.5 \\
CWP & 51.0 & 69.7 & 61.1 & 81.7 & 21.2 & 50.9 \\
ArcFace~\cite{deng2019arcface}\dag & 7.9 & 29.4 & 20.2 & 45.0 & 19.3 & 57.6 \\
SupCon~\cite{khosla2020supervised}\dag & 11.4 & 43.1 & 16.7 & 42.9 & 14.5 & 56.9 \\
StableRep+~\cite{tian2023stablerep}\dag & 62.0 & 89.1 & 22.1 & 55.8 & 25.3 & 67.0 \\
MaPLe~\cite{khattakMaPLe} & 64.7 & 85.6 & 86.7 & 96.8 & 36.9 & 74.6 \\
LoRA~\cite{hu2022lora} & 72.3 & 90.6 & 87.6 & 96.8 & 28.9 & 64.0 \\
\textbf{Ours-Query} & \textbf{83.8} & \textbf{93.5} & \textbf{98.7} & \textbf{99.8} & \textbf{51.0} & \textbf{78.1} \\
\bottomrule
\end{tabular}

\end{table}
\noindent\textbf{Additional Retrieval Metrics.}
To further validate retrieval behavior beyond mean Average Precision, we report Recall@1 and Recall@5 in Table~\ref{tab:recall}. These metrics highlight how early each method surfaces an attribute-consistent result when retrieving from a frozen gallery. Across all datasets, our text-conditioned transformation maintains strong early-rank performance, reliably retrieving an attribute-matching image within the first few items.

\begin{table}
\centering
\caption{Mean average precision (mAP) retrieval performance across all benchmarks allowing for transformed galleries. Scores are averaged across taxonomies within each dataset. Our method remains competitive with attribute-specific trained models. Memory footprint details explained in Section~\ref{sec:imp_details}.}
\setlength{\tabcolsep}{1pt}
\label{tab:global_ret}
\begin{tabular}{lcccc} 
\toprule
Method & \begin{tabular}[c]{@{}c@{}}Memory\\Footprint\end{tabular} & \begin{tabular}[c]{@{}c@{}}Clevr4\\mAP\end{tabular} & \begin{tabular}[c]{@{}c@{}}Stanford40\\mAP\end{tabular} & \begin{tabular}[c]{@{}c@{}}ShotBench\\mAP\end{tabular} \\ 
\hline
Base Features & 6GB & 33.8 & 43.7 & 25.2 \\
LCR & 84GB & 52.8 & 46.7 & 22.0 \\
CWP & 84GB & 55.6 & 53.2 & 23.2 \\
ArcFace~\cite{deng2019arcface}\dag & 84GB & 86.4 & 61.1 & 38.8 \\
SupCon~\cite{khosla2020supervised}\dag & 84GB & \textbf{89.1} & 66.3 & 42.6 \\
StableRep+~\cite{tian2023stablerep}\dag & 84GB & 88.8 & 60.8 & \textbf{43.1} \\
MaPLe~\cite{khattakMaPLe} & 84GB & 50.4 & 48.6 & 23.9 \\
LoRA~\cite{hu2022lora} & 84GB & 79.9 & 54.8 & 25.1 \\
\textbf{Ours-Global} & 6GB & 85.6 & \textbf{95.7} & 41.9 \\
\bottomrule
\end{tabular}
\end{table}
\begin{table*}[ht]
\centering
\caption{Category-wise clustering accuracy results across multiple datasets. Category names are abbreviated (full names in Section~\ref{sec:exp_details}). Highlighted columns denote averages across taxonomies. Our approach achieves strong, consistent performance across both semantic and perceptual visual attributes.}
\label{tab:multiple_clustering}
\vspace{-2mm}
\setlength{\tabcolsep}{2.5pt}
\resizebox{\linewidth}{!}{
\begin{tabular}{lccccc|cccc|cccccccc} 
\toprule
 & \multicolumn{5}{c|}{Clevr-4} & \multicolumn{4}{c|}{Stanford 40 Action} & \multicolumn{8}{c}{ShotBench} \\
Method & Col & Cnt & Shp & Tex & Avg. & Act & Loc & Md & Avg. & CA & CM & LS & LI & LT & SF & SS & Avg. \\ 
\midrule
Base Features & 13.1 & 13.1 & 89.2 & 13.9 & {\cellcolor[rgb]{0.855,0.914,0.973}}32.3 & 59.4 & 58.6 & \underline{75.8} & {\cellcolor[rgb]{0.855,0.914,0.973}}64.6 & \underline{51.9} & \underline{28.2} & \underline{38.5} & 23.3 & 26.8 & 30.3 & 28.4 & {\cellcolor[rgb]{0.855,0.914,0.973}}32.5 \\
LCR & \textbf{85.9} & 28.3 & 81.9 & 29.4 & {\cellcolor[rgb]{0.855,0.914,0.973}}56.4 & 64.8 & 65.5 & 64.2 & {\cellcolor[rgb]{0.855,0.914,0.973}}64.8 & 35.8 & 24.2 & 37.5 & \underline{25.8} & 25.5 & 27.8 & \underline{34.8} & {\cellcolor[rgb]{0.855,0.914,0.973}}30.2 \\
CWP & 80.2 & 35.6 & \textbf{97.3} & 28.0 & {\cellcolor[rgb]{0.855,0.914,0.973}}60.3 & 74.8 & 59.1 & 60.8 & {\cellcolor[rgb]{0.855,0.914,0.973}}64.9 & 37.2 & 27.1 & 38.3 & \underline{25.8} & 29.8 & \underline{33.9} & 33.4 & {\cellcolor[rgb]{0.855,0.914,0.973}}32.2 \\ 
Multi-Map\venue{CVPR'24} & 75.3 & 53.9 & 65.5 & 56.5 & {\cellcolor[rgb]{0.855,0.914,0.973}}62.8 & 57.9 & 59.4 & 71.0 & {\cellcolor[rgb]{0.855,0.914,0.973}}62.8 & 45.1 & 28.0 & 38.2 & 23.8 & \underline{31.0} & 29.9 & 31.9 & {\cellcolor[rgb]{0.855,0.914,0.973}}\underline{32.6} \\
Multi-Sub\venue{NeurIPS'25} & \underline{84.7} & \underline{63.3} & 74.9 & \underline{65.9} & {\cellcolor[rgb]{0.855,0.914,0.973}}\underline{72.2} & 60.4 & 63.1 & 74.7 & {\cellcolor[rgb]{0.855,0.914,0.973}}66.1 & 44.0 & 28.0 & 37.9 & 24.0 & 29.9 & 31.4 & 31.3 & {\cellcolor[rgb]{0.855,0.914,0.973}}32.4 \\
IC$|$TC\venue{ICLR'24} & 53.4 & 43.5 & 71.9 & 62.8 & {\cellcolor[rgb]{0.855,0.914,0.973}}57.9 & 77.7 & \textbf{75.0} & 75.5 & {\cellcolor[rgb]{0.855,0.914,0.973}}\underline{76.1} & 27.5 & 18.0 & 31.6 & 20.3 & 13.7 & 18.0 & 23.1 & {\cellcolor[rgb]{0.855,0.914,0.973}}21.7 \\
SSD-LLM\venue{ECCV'24} & 49.1 & 44.2 & 72.0 & 60.8 & {\cellcolor[rgb]{0.855,0.914,0.973}}56.5 & 81.4 & \underline{70.4} & 70.6 & {\cellcolor[rgb]{0.855,0.914,0.973}}74.1 & 26.7 & 23.5 & 29.1 & 23.2 & 20.1 & 29.0 & 23.4 & {\cellcolor[rgb]{0.855,0.914,0.973}}25.0 \\
X-Cluster\venue{arxiv'25} & 70.3 & \textbf{65.7} & 58.4 & 65.3 & {\cellcolor[rgb]{0.855,0.914,0.973}}64.9 & \underline{82.8} & 69.8 & 52.3 & {\cellcolor[rgb]{0.855,0.914,0.973}}68.3 & - & - & - & - & - & - & - & {\cellcolor[rgb]{0.855,0.914,0.973}}- \\ 
\textbf{Ours-Global} & 72.1 & 59.0 & \underline{87.6} & \textbf{73.5} & {\cellcolor[rgb]{0.855,0.914,0.973}}\textbf{73.0} & \textbf{84.4} & 67.3 & \textbf{88.3} & {\cellcolor[rgb]{0.855,0.914,0.973}}\textbf{80.0} & \textbf{65.9} & \textbf{32.6} & \textbf{61.9} & \textbf{25.9} & \textbf{38.9} & \textbf{74.6} & \textbf{61.8} & {\cellcolor[rgb]{0.855,0.914,0.973}}\textbf{51.6} \\
\bottomrule
\end{tabular}
}
\end{table*}
\begin{table}
\centering
\caption{Multiple clustering performance across more benchmarks, averaged across taxonomies within each dataset. Our method strongly improves clustering accuracy.}
\label{tab:extra_mc}
\setlength{\tabcolsep}{3.0pt}
\small
\begin{tabular}{lcc|c|cc|c} 
\toprule
\multirow{2}{*}{Method} & \multicolumn{2}{c|}{Cards} & FER & \multicolumn{2}{c|}{Fruits} & \multirow{2}{*}{\begin{tabular}[c]{@{}c@{}}All\\Avg.\end{tabular}} \\
 & \multicolumn{1}{l}{Rank} & \multicolumn{1}{l|}{Suit} & \multicolumn{1}{l|}{Emotion} & \multicolumn{1}{l}{Fruit} & \multicolumn{1}{l|}{Style} &  \\ 
\midrule
Base Features & 30.4 & 63.1 & 27.0 & 52.7 & 33.1 & {\cellcolor[rgb]{0.855,0.914,0.973}}38.9 \\
LCR & 56.5 & 93.5 & 23.0 & \textbf{86.4} & 86.5 & {\cellcolor[rgb]{0.855,0.914,0.973}}61.5 \\
CWP & 71.9 & 92.3 & 23.9 & 71.3 & 80.8 & {\cellcolor[rgb]{0.855,0.914,0.973}}60.7 \\
\textbf{Ours-Global} & \textbf{78.5} & \textbf{99.2} & \textbf{65.7} & 85.3 & \textbf{100.0} & {\cellcolor[rgb]{0.855,0.914,0.973}}\textbf{82.4} \\
\bottomrule
\end{tabular}
\end{table}

\noindent\textbf{Modified Gallery Attribute-aware Retrieval.} In this setting, we remove the frozen-gallery constraint and allow each method to transform both the query and the gallery embeddings. Although this setting is less practical than the query mode, it provides a direct comparison to approaches that rely on full-space adaptation. As shown in Table~\ref{tab:global_ret}, our learned transform, despite being trained for a query transformation setting, remains competitive with methods that train separate attribute-specific adapters. Importantly, these adapter-based approaches require one model per attribute and a corresponding recomputation of gallery embeddings, leading to substantially higher complexity and memory that scale linearly with the number of attributes.

\noindent\textbf{Expanded Multiple-clustering Results.}
To complement the averaged clustering results reported in the main paper, we provide per-attribute accuracies for all datasets in Table~\ref{tab:multiple_clustering}. These finer-grained scores reveal how performance varies across individual attributes, offering a clear view of attribute difficulty and the consistency of the learned transformation. Across nearly all attributes, the global transform maintains stable clustering performance.

\noindent\textbf{Additional Datasets: Multiple-clustering.}
We additionally evaluate clustering on Card, FER, and FruitSALAD to demonstrate broad attribute coverage in Table~\ref{tab:extra_mc}. In all three benchmarks, the global transformation produces stable and competitive clustering accuracy, confirming that the same attribute-aware transform generalizes effectively across diverse visual domains.

\noindent\textbf{Conditional Similarity Networks Comparison.} Conditional Similarity Networks (CSN)~\cite{veit2017conditional} trains attribute masks while tuning ResNet-18, not satisfying our frozen-gallery constraint. We compare against CSN's published performance on Zappos50k~\cite{veit2017conditional} and against an implementation of their method with a frozen-CLIP-backbone. Table~\ref{tab:csn} shows our method wins each comparison, with the widest margins on the more semantic Stanford-40 dataset.

\begin{table}[h]
\centering
\footnotesize
\setlength{\tabcolsep}{5pt}
\renewcommand{\arraystretch}{0.85}
\caption{Comparison to Conditional Similarity Networks adapted baseline. Our method wins each comparison.}
\label{tab:csn}
\begin{tabular}{lccc}
\toprule
Dataset & \makecell{CSN (reported)\\w/ResNet-18} & \makecell{CSN (our impl.)\\w/ CLIP} & Ours \\
\midrule
Clevr-4 (mAP)         & ---  & 70.6 & \textbf{77.6} \\
Stanford-40 (mAP)     & ---  & 46.5 & \textbf{84.4} \\
Zappos (triplet acc.) & 78.8 & 80.9 & \textbf{85.4} \\
\bottomrule
\end{tabular}
\end{table}

\noindent\textbf{Hyperparameter Sensitivity Analysis.}
We evaluate the effect of the multimodal guidance weight $\omega$ on retrieval and clustering performance in \Cref{tab:hyperparameter_analysis}. Moderate guidance improves retrieval while maintaining similar clustering performance, with $\omega=0.1$ performing best overall and therefore used in our experiments. As $\omega$ increases further, clustering accuracy improves, while retrieval performance begins to degrade. This behavior suggests that stronger text alignment increasingly reshapes the embedding geometry, which can benefit clustering structure but reduce the fine-grained similarity relationships needed for retrieval.

\begin{table}[h]
\caption{Effect of multimodal guidance weight $\omega$ on CLEVR-4 retrieval (mAP) and clustering accuracy. Moderate guidance provides the best overall performance, while larger values slightly improve clustering at the expense of retrieval.}
\label{tab:hyperparameter_analysis}
\centering
\begin{tabular}{lcc} 
\toprule
Loss Weight & Retrieval & Clustering  \\ 
\midrule
$\omega = 0$                      & 72.3      & 78.8        \\
$\omega = 0.01$                   & 72.4      & 79.6        \\
$\omega = 0.1$                    & 73.4      & 79.6        \\
$\omega = 1$                      & 73.0      & 81.5        \\
\bottomrule
\end{tabular}
\end{table}

\begin{wraptable}{r}{0.38\linewidth}
\vspace{-3.5mm}
\centering
\caption{Zero-shot classification and k-NN evaluation (k=5) on CIFAR10. The original class names are not captured in unrelated "color" transformation, yet k-NN evaluation shows that \textit{relative} similarities are preserved.}
\small
\begin{tabular}{@{}lll@{}}
\toprule
               & \multicolumn{2}{c}{CIFAR10}                                            \\
Method         & \multicolumn{1}{c}{ZS Acc.} & \multicolumn{1}{c}{k-NN} \\ \midrule
Base Features       & 86.06                                 & 97.66                          \\
Ours + "color" & 52.57                                  & 97.48                          \\ \bottomrule
\end{tabular}
\label{tab:cifar_knn}
\vspace{-6mm}
\end{wraptable}

\noindent\textbf{Does the transformation preserve general visual structure beyond the target attribute?}
Although the transformed embeddings primarily emphasize the desired attribute, we find that meaningful global relationships remain intact. Table~\ref{tab:cifar_knn} shows that our transformation does not collapse or distort the underlying feature space merely to satisfy the contrastive loss. Text-based classification for unseen class names drops only slightly, and k-NN evaluations confirm that relative similarity is preserved—samples that are visually and semantically similar remain close. For example, a “blue airplane” remains closer to other blue airplanes than to blue cars, although it is closer to both compared to a "yellow airplane". This indicates that the transformation sharpens attribute-specific structure without erasing general semantics.

\noindent\textbf{Cross-dataset generalization on MS-COCO.}
To further evaluate generalization, we test the trained model on images from the MS-COCO test2017 split without any additional training. Since COCO does not contain the attribute taxonomies used in our experiments, we automatically annotate images with the relevant attributes and subclasses using Qwen3-VL-32B with a structured prompting scheme. The trained hypernetwork remains unchanged and is applied directly to the COCO embeddings.

\Cref{tab:coco_generalization} reports attribute-aware retrieval performance under this setting. Despite the relatively small size of the training datasets and a notable distribution shift between the training datasets and COCO (e.g., Clevr-4 provides synthetic object images for learning color and count attributes), our method consistently improves over the baseline across all attribute categories. These results suggest that the learned transformations capture transferable attribute structure encoded in the foundation model embeddings rather than overfitting to the specific training datasets.

\begin{table}
\centering
\caption{Zero-shot attribute retrieval performance on MS-COCO. A model trained only on the source datasets (CLEVR-4, Stanford40, and ShotBench) is evaluated on COCO images annotated with attribute labels for evaluation. Results show consistent improvements across all attributes, indicating that the learned transformations capture transferable attribute structure beyond the training distributions.}
\label{tab:coco_generalization}
\begin{tabular}{lccccccc} 
\toprule
Method                & Action        & Location      & Mood          & Color         & Count         & Camera Angle  & Shot Size      \\ 
\hline
Baseline              & 21.9          & 36.6          & 34.9          & 19.7          & 30.5          & 36.4          & 30.7           \\
\textbf{Ours - Query} & \textbf{27.4} & \textbf{47.1} & \textbf{37.4} & \textbf{24.7} & \textbf{34.8} & \textbf{42.0} & \textbf{39.3}  \\
\bottomrule
\end{tabular}
\end{table}

\begin{figure*}
    \centering
    \includegraphics[width=.9\linewidth]{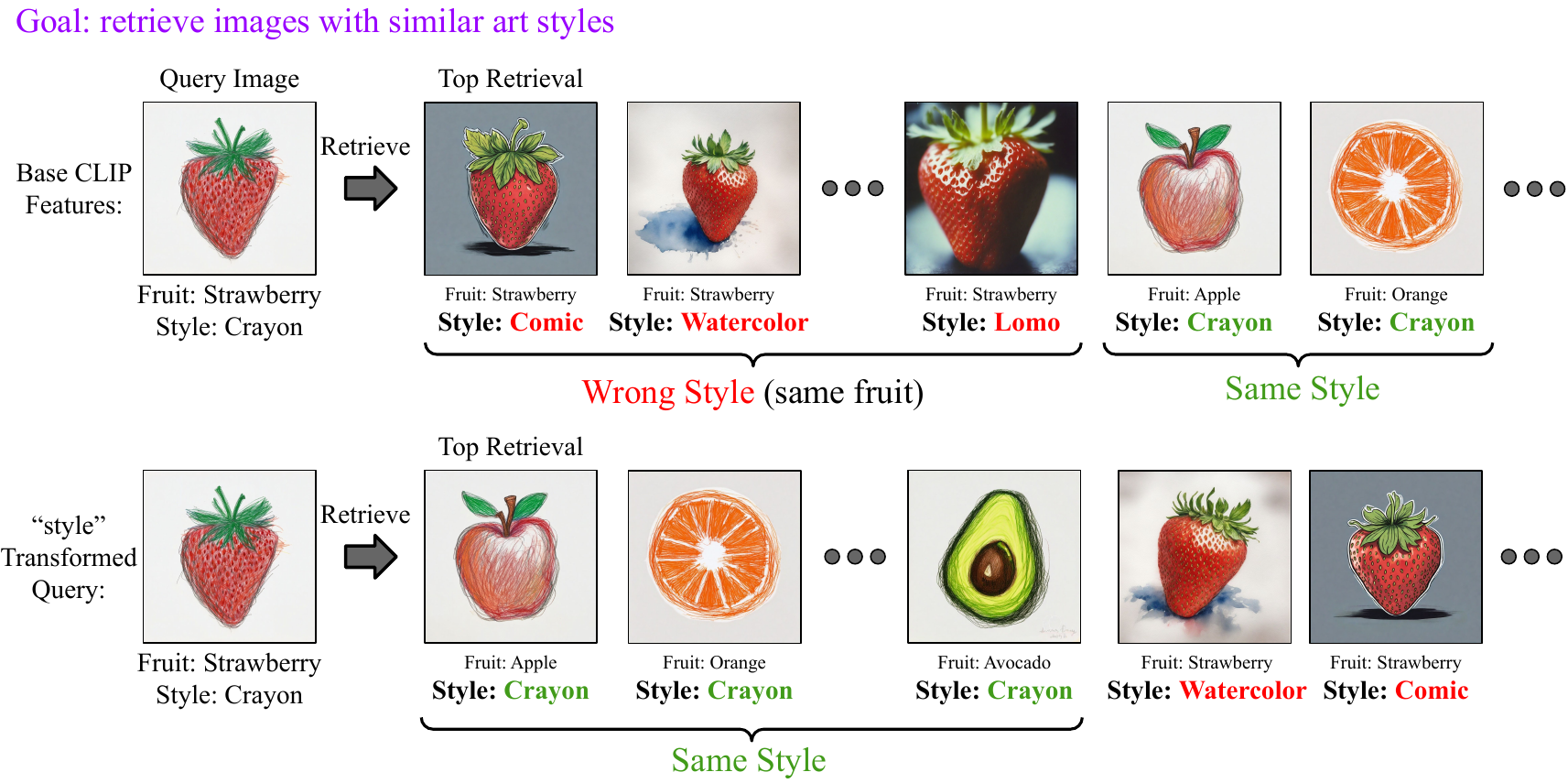}
    \caption{Retrieval ranked by similarity using base CLIP embeddings and our transformed query (“style”).  CLIP prioritizes object semantics (e.g., fruit type), whereas our transformation maintains attribute alignment—retrieving different fruits rendered in the same artistic style.}
    \label{fig:qual_fruit}
\end{figure*}

\begin{figure*}
    \centering
    \includegraphics[width=.95\linewidth]{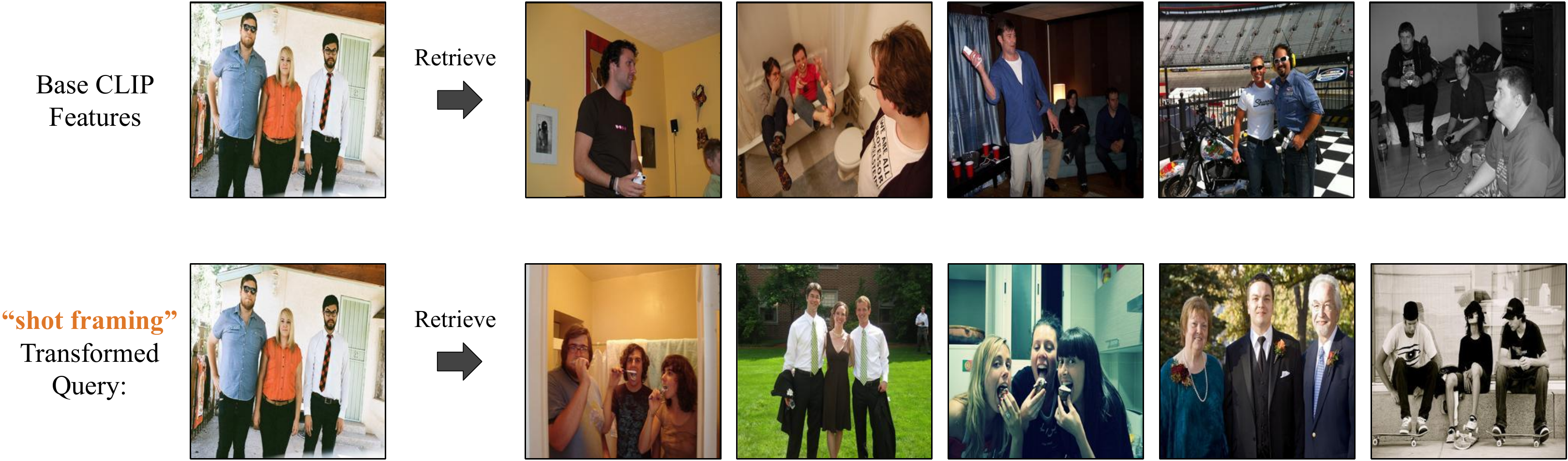}
    \caption{Top retrieval results comparing CLIP embeddings and our text-guided transformation using the query “shot framing.” Our method surfaces images with matching three-subject shot-framing structure.}
    \label{fig:qual_shotframe}
\end{figure*}

\begin{figure*}
    \centering
    \includegraphics[width=.95\linewidth]{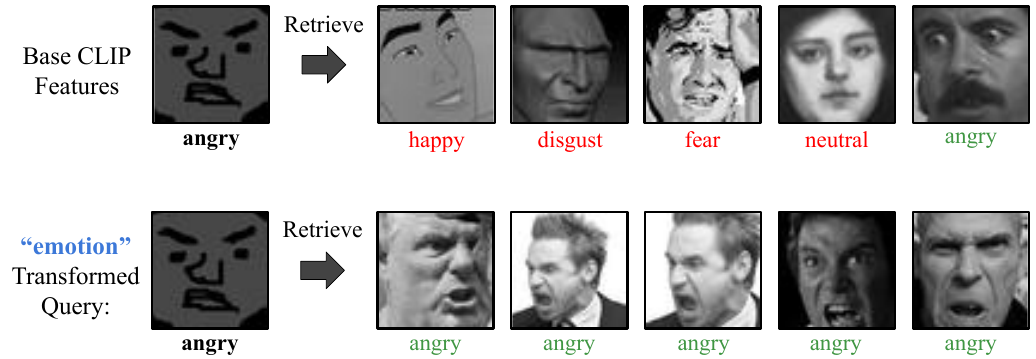}
    \caption{Attribute-aware retrieval using the “emotion” query. Despite the input being a stylized drawing, the learned transform isolates the intended emotion and retrieves aligned “angry” expressions across diverse visual styles.}
    \label{fig:qual_emotion}
\end{figure*}

\begin{figure*}
    \centering
    \includegraphics[width=.95\linewidth]{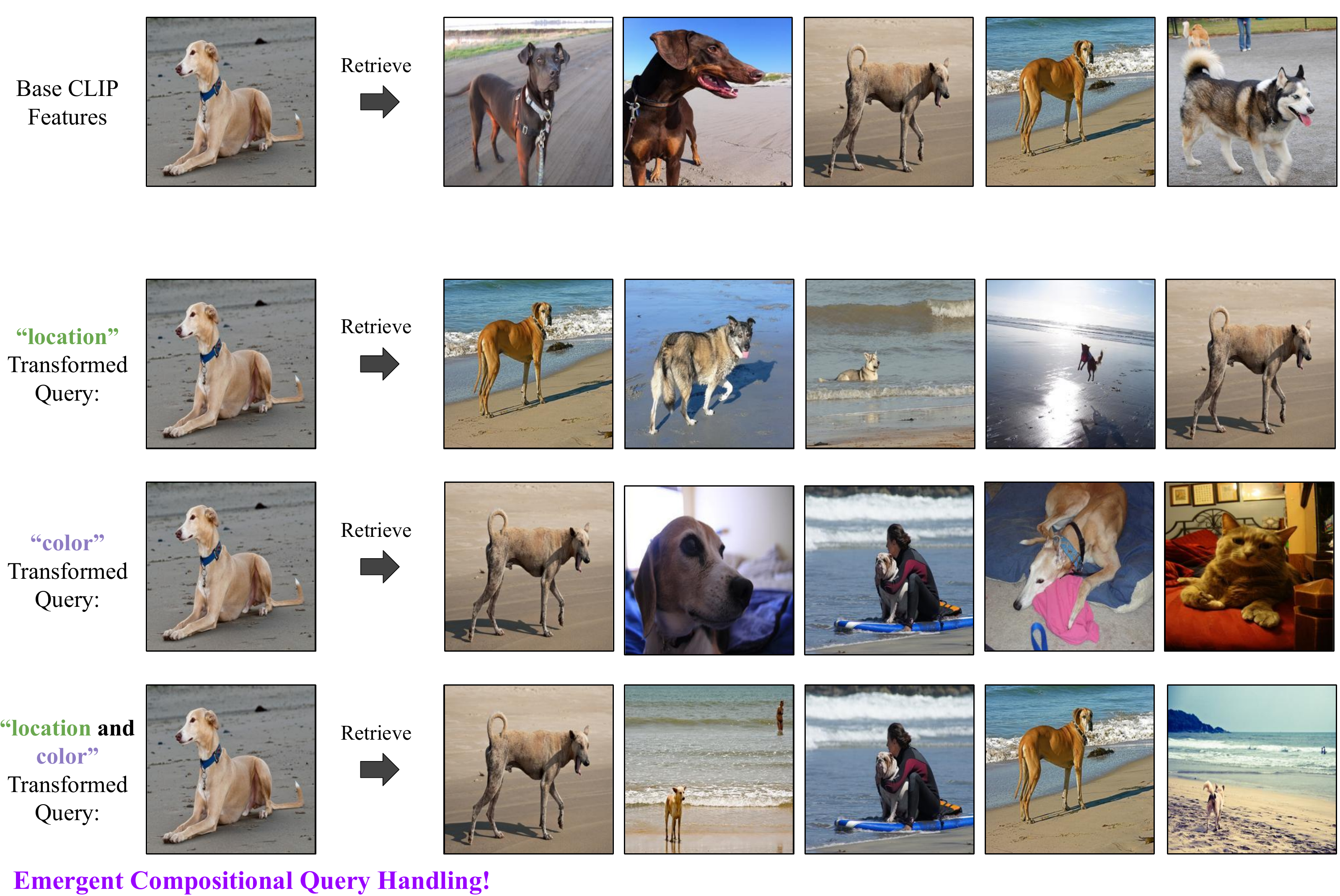}
    \caption{Retrieval comparison illustrating compositional attribute control. CLIP primarily retrieves dog-related images, whereas our transformations isolate specific attributes: the “color” transform finds dogs of similar color, the “location” transform retrieves dogs on beaches, and the combined query “location and color” retrieves beach scenes with dogs of similar coloration, demonstrating emergent multi-attribute controllability.}
    \label{fig:qual_compositional}
\end{figure*}

\noindent\textbf{Qualitative Results.} Beyond quantitative evaluations, we visualize the behavior of our text-conditioned transformations across a range of attributes and visual domains. Figure~\ref{fig:qual_fruit} visualizes ranked retrieval results on the Fruit-SALAD dataset using the “style” attribute. Our transform causes the retrieval to respect the input attribute first, retrieving different fruit images in the same style before retrieving same fruit images in a different style. Figure \ref{fig:qual_shotframe} highlights retrievals driven by the shot framing attribute, where our method prioritizes compositional structure over object identity. Figure \ref{fig:qual_emotion} demonstrates robustness to stylistic variation: even with a stylized drawing as input, the “emotion” transform retrieves semantically aligned expressions. Finally, Figure \ref{fig:qual_compositional} illustrates emergent compositional control by showing that color, location, and their combination produce distinct and semantically coherent retrieval sets. Collectively, these examples show that the learned transformations cleanly isolate visual attributes, generalize across domains, and support multi-attribute control.

\end{document}